%% file: root.tex
\documentclass[letterpaper, 10 pt, journal, twoside]{IEEEtran} 

\IEEEoverridecommandlockouts                              

\makeatletter
\let\NAT@parse\undefined
\makeatother
\usepackage[numbers,sort&compress]{natbib}

\usepackage[pdftex]{graphicx}
\usepackage{amsmath}
\usepackage{amssymb}  
\usepackage{subfigure}
\usepackage{subcaption}
\usepackage{multirow}
\usepackage{array,booktabs}
\usepackage{diagbox}
\usepackage{balance}
\usepackage{xspace}
\usepackage{algorithm}
\usepackage{algpseudocode}
\usepackage{adjustbox}

\usepackage[final]{hyperref}
\hypersetup{
 colorlinks=true,
 linkcolor=magenta,
 filecolor=magenta,
 urlcolor=magenta,
 citecolor=black
}
\usepackage{cleveref}
\crefname{figure}{Fig.}{Figs.}
\Crefname{figure}{Fig.}{Figs.}
\usepackage[font=small]{caption}
\usepackage{rpm_SIunits}
\usepackage{rpm_acronyms}
\usepackage{rpm_math}
\usepackage{rpm_misc}
\usepackage{bbm}

\usepackage{textcomp}

\usepackage{soul,color}
\usepackage{lipsum}
\usepackage{dsfont}

\usepackage{xcolor}
\usepackage{colortbl}

\usepackage{amssymb}
\usepackage{pifont}
\usepackage{tikz} 
\usepackage{subcaption}
\usepackage{caption}
\newlength{\savedtextbaselineskip}

\makeatletter
\renewcommand{\section}{%
  \@startsection{section}{1}{\z@}%
    {2ex plus 0.5ex minus 0.5ex}
    {0.7ex plus 1ex}
    {\normalfont\normalsize\centering\scshape}%
}
\makeatother

\makeatletter
\renewcommand{\subsection}{%
  \@startsection{subsection}{2}{\z@}%
    {1.5ex plus 0.2ex minus 0.2ex}
    {0.4ex plus 0.1ex}
    {\normalfont\normalsize\itshape}%
}
\makeatother

\title{\fontsize{16.5}{18}\selectfont\bfseries RAGNAROK: Radar-Aided Gravity-Normalized Alignment for Robust Open Keyframe-based Radar-Visual-Kinematic-Inertial SLAM}

\author{Hanjun Kim$^{1}$, Chiyun Noh$^{1}$, Sangwoo Jung$^{1}$, Jaehyung Jung$^{2}$,\\
Simon Boche$^{2}$, Cedric Le Gentil$^{3}$, Stefan Leutenegger$^{3}$ and Ayoung Kim$^{1*}$

\thanks{Manuscript received: June 16, 2026; Revised September 10, 2026; Accepted October 7, 2026.}
\thanks{This paper was recommended for publication by Editor Javier Civera upon evaluation of the Associate Editor and Reviewers’ comments.
This work was supported by the NRF grant (No. RS-2023-00241758).}
\thanks{$^{1}$H. Kim, C. Noh, S. Jung and A. Kim are with the Dept. of Mechanical Engineering, SNU, Seoul, S. Korea {\tt\small [hanjun815, gch06208, dan0130, ayoungk]@snu.ac.kr}}
\thanks{$^{2}$J. Jung and S. Boche are with the Mobile Robotics Lab, School of Computation, Information and Technology (CIT) at the Technical University of Munich (TUM), 80333 Munich, Germany, and with the Munich Institute of Robotics and Machine Intelligence (MIRMI). {\tt\small [jaehyung.jung, simon.boche]@tum.de}}
\thanks{$^{3}$C. Le Gentil and S. Leutenegger are with the Mobile Robotics Lab, ETH Zurich, 80892 Zurich, Switzerland {\tt\small [clegentil, lestefan]@ethz.ch}}
\thanks{Digital Object Identifier (DOI): see top of this page.}
}
\IEEEaftertitletext{\vspace*{-6mm}}
\begin{document}

\maketitle

\input{0_abstract}

\input{1_introduction}
\input{3_method}
\input{4_experiment}
\input{5_conclusion}
\input{6_acknowledgments}

\balance
\small
\bibliographystyle{IEEEtranN} 
\bibliography{string-short,references}

\end{document}

%% file: 0_abstract.tex
\begin{abstract}
Legged robots offer superior mobility in unstructured environments, but reliable operation in such conditions requires robust state estimation. 
To address the vulnerability of proprioceptive estimators in rough terrain, recent methods have incorporated radar to provide velocity measurements.
However, their limited yaw observability still leads to drift, and failure-aware fusion for adverse environments remains underexplored. 
In this letter, we present \textbf{RAGNAROK}, the first radar-visual-kinematic-inertial SLAM designed for robust operation in challenging environments.
It integrates slip- and rolling-contact-aware leg velocity estimation, a kinematics-aware radar factor, and degradation-aware image enhancement. 
We further incorporate a B-spline-based radar-aided proprioceptive backbone, adaptive weighting, and online extrinsic calibration.
Extensive experiments on public and self-collected datasets demonstrate that RAGNAROK achieves robust performance under challenging conditions and outperforms state-of-the-art baselines.
The source code and dataset are available at \href{https://github.com/hanjun815/RAGNAROK}{https://github.com/hanjun815/RAGNAROK}.
\end{abstract}

%% file: 1_introduction.tex
\section{Introduction}
\label{sec:intro}

\IEEEPARstart{L}{egged} robots have attracted increasing attention for real-world applications such as inspection, exploration, and search-and-rescue, due to their ability to traverse uneven surfaces and unstructured terrain \cite{tranzatto2022cerberus}.
Reliable operation in these scenarios requires accurate state estimation, since precise localization is fundamental for safe and efficient motion planning. Early studies primarily relied on proprioceptive sensing, typically fusing leg joint encoders and an \ac{IMU} \cite{bloesch2012state}.
However, such approaches remain vulnerable to unmodeled physical interactions, including contact impacts, vibrations, and foot slippage. 

To overcome these limitations, radar has emerged as a compelling complementary sensing modality in recent studies.
In particular, \ac{FMCW} radar provides radial velocity measurements of surrounding points through frequency modulation, enabling direct ego-velocity estimation. 
Early efforts integrated these radar velocities with leg kinematic factors;
however, vertical drift remained because roll and pitch estimation still depended heavily on gyroscope integration \cite{jung2024co}. Subsequent work addressed this issue by introducing continuous-time local gravity estimation, thereby improving roll and pitch estimation and reducing drift along the vertical axis \cite{noh2026garlileo}.

\begin{figure}[!t]
    \centering
    \includegraphics[trim= 8.4cm 4.6cm 8.4cm 4.8cm, clip,width=\columnwidth]{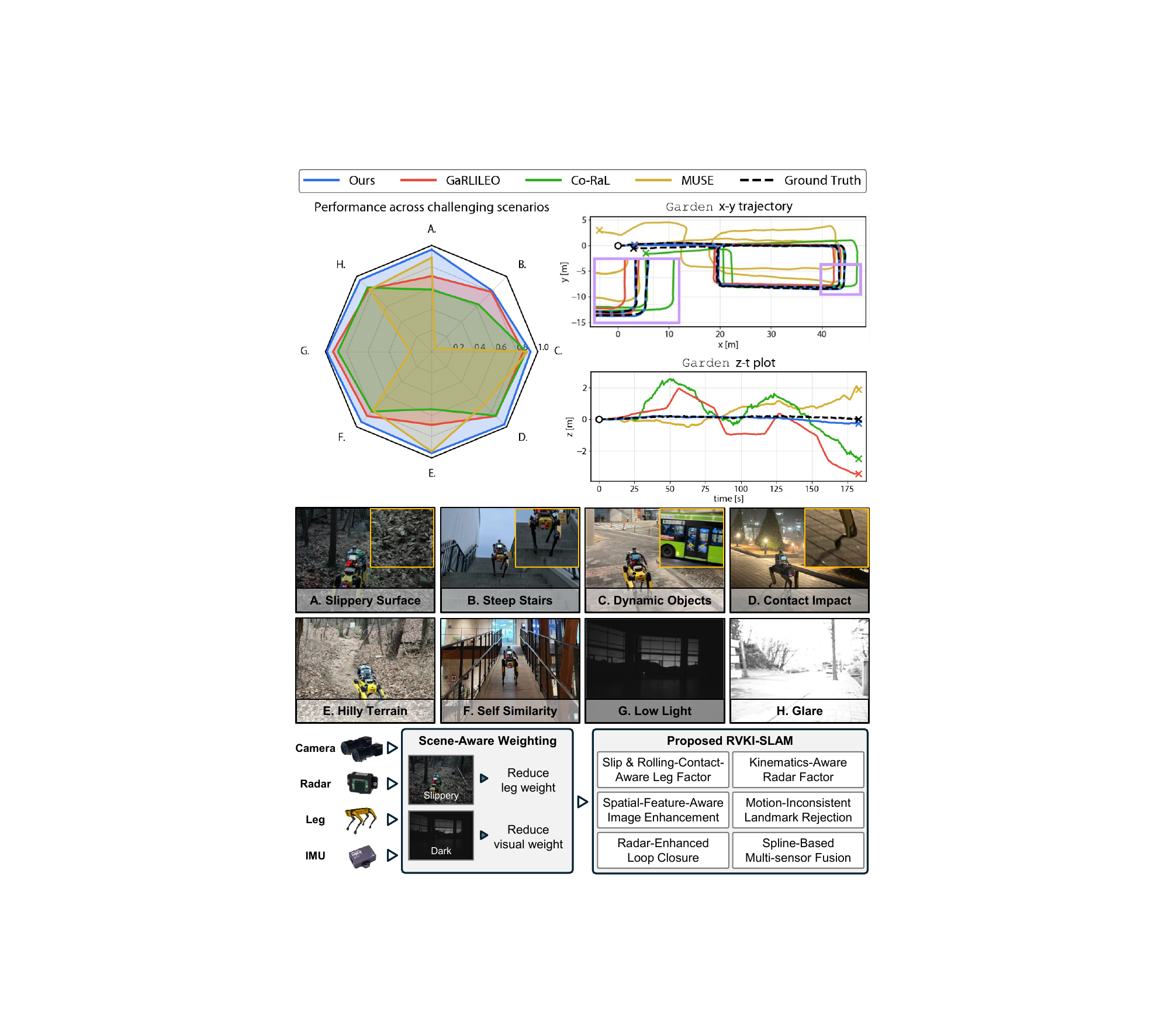}
    \vspace{-5mm}
    \caption{\textbf{Top:} Performance comparison with SOTA baselines, with scores computed as $\mathrm{ATE}_{\mathrm{mean}} / (\mathrm{ATE}_{\mathrm{mean}} + \mathrm{ATE})$. \textbf{Middle:} Representative challenging scenarios from the self-collected dataset. \textbf{Bottom:} Overview of the proposed RAGNAROK pipeline.}
    \label{fig:Fig1}
    \vspace{-6mm}
\end{figure}

Despite this progress, several limitations remain. First, limited constraints on relative yaw orientation, as gravity constrains roll and pitch but not yaw~\cite{kubelka2022gravity}, cause inevitable horizontal drift over long trajectories.
Second, sparse
radar measurements alone are insufficient to support loop closure and mapping, limiting their extension to a robust \ac{SLAM} system. These limitations can be mitigated by integrating visual information, which offers dense textural and semantic cues. 
However, seamless multimodal fusion remains challenging because each sensing modality exhibits distinct failure modes. 
Robust state estimation in extreme environments therefore requires explicit modeling of modality-specific failures, including visual degradation and kinematic errors. 
Nevertheless, visual measurements are often incorporated without degradation-aware active enhancement.
Existing multi-sensor fusion methods for legged robots either discard unreliable measurements or rely on fixed weighting schemes \cite{camurri2020pronto,nistico2025muse}, limiting the full exploitation of exteroceptive information.
In particular, failure-aware visual fusion integrated with radar-aided gravity estimation for robust roll and pitch constraints remains largely unexplored.

To bridge this gap, we propose \textbf{RAGNAROK}, \underline{\textbf{R}}adar-\underline{\textbf{A}}ided \underline{\textbf{G}}ravity-\underline{\textbf{N}}ormalized \underline{\textbf{A}}lignment for \underline{\textbf{R}}obust \underline{\textbf{O}}pen-\underline{\textbf{K}}eyframe-based Radar-Visual-Kinematic-Inertial \ac{SLAM}.
As illustrated in \figref{fig:Fig1}, it achieves robust performance in challenging environments by combining degeneracy-aware enhancement with adaptive sensor weighting.
It leverages a slip- and rolling-contact-aware leg factor and a kinematics-aware multi-radar factor, jointly optimized with local gravity estimation. 
To mitigate yaw drift and preserve loop-closure capability under perceptual degradation, we incorporate spatial-feature-aware image enhancement and radar-enhanced place recognition, respectively.
A continuous-time radar-aided proprioceptive fusion backbone interpolates motion between camera frames, allowing velocity and gravity cues to be propagated to both the visual frontend and backend.
Finally, we release the first open-source legged-robot dataset combining measurements from multiple radars and RGB cameras, a stereo \ac{IR} camera, LiDAR, proprioceptive sensors, and an \ac{IMU}.

Our main contributions are summarized as follows:
\begin{itemize}
    \item We present RAGNAROK, a novel exteroceptive-proprioceptive \ac{SLAM} framework with degeneracy-aware adaptive fusion. To the best of our knowledge, this is the first radar-visual-kinematic-inertial \ac{SLAM} designed for robust operation in challenging environments.

    \item We propose a robust leg factor that accounts for two slip modes and rolling contact, enabling reliable estimation over rough terrain. In addition, our kinematics-aware orthogonal radar factor dynamically weights Doppler residuals according to radar placement and motion.

    \item We introduce a novel spatial-feature-aware image enhancement method that enables dense and spatially uniform feature extraction. 
    Moreover, our proprioception-aided visual frontend aligns feature descriptors with local gravity and rejects motion-inconsistent landmarks.
    
    \item We validate RAGNAROK against \ac{SOTA} methods across diverse conditions (outdoor, low-light, glare, and slippage), and demonstrate superior performance in challenging environments (forested terrain, stair traversal, and self-similar corridors). 
\end{itemize}

%% file: 3_method.tex
\begin{figure*}[!t]
    \centering
    \includegraphics[trim= 2.4cm 1.3cm 2.3cm 1.3cm, clip,width=0.95\textwidth]{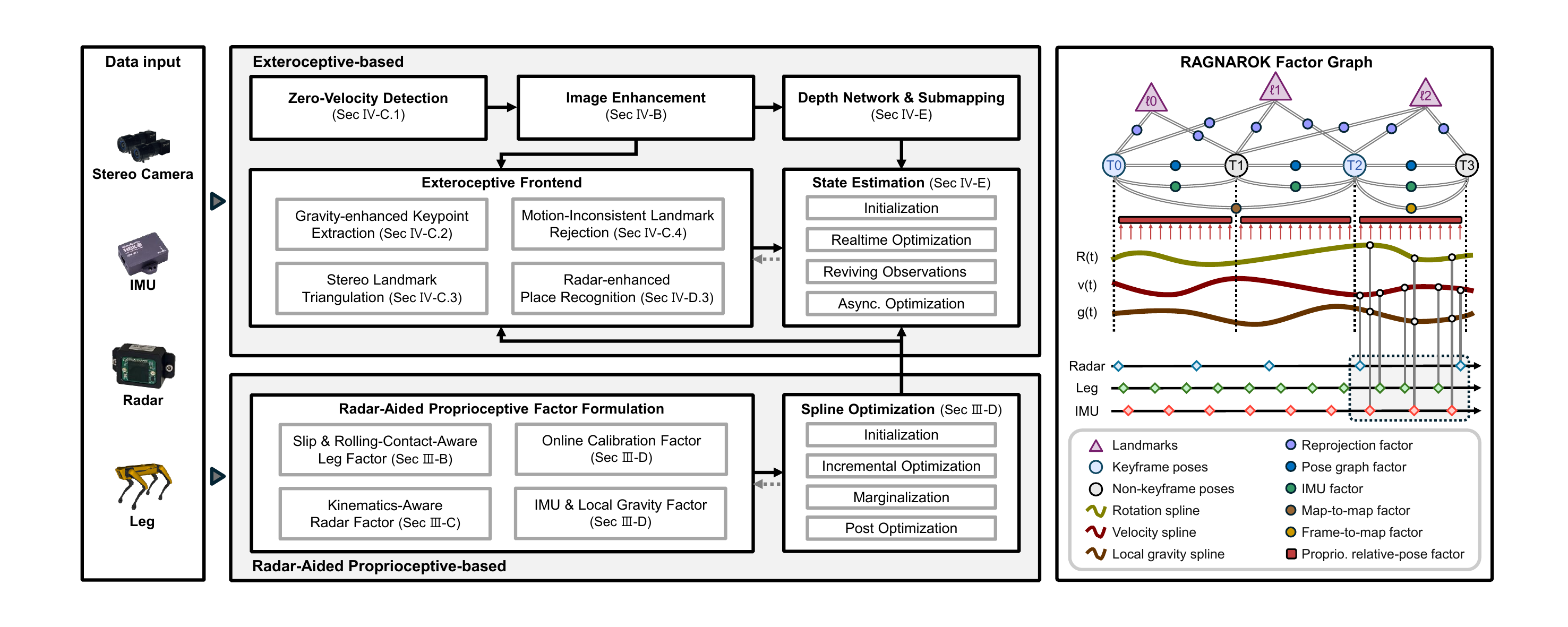}
    \vspace{-2mm}
    \caption{Overview of RAGNAROK:
    The radar-aided proprioceptive estimator optimizes the splines of velocity, rotation, and local gravity.
    The exteroceptive estimator performs image enhancement, keypoint matching, landmark triangulation, place recognition, and submapping.}
    \label{fig:overview}
    \vspace{-5mm}
\end{figure*}

\section{Proprioceptive-based State Estimator}
As illustrated in \figref{fig:overview}, the proposed system consists of a radar-aided proprioceptive state estimator and an exteroceptive state estimator. Our radar-aided proprioceptive estimator extends the gravity-aligned continuous-time estimation of GaRLILEO \cite{noh2026garlileo} to rough-terrain contacts and kinematics-aware multi-radar constraints.
It optimizes B-spline representations of ego-centric velocity, rotation, and local gravity using a novel slip- and rolling-contact-aware leg factor and a kinematics-aware multi-radar factor. 
The resulting spline is queried at visual keyframe timestamps and used by the exteroceptive estimator as relative motion constraints.

\subsection{State Definition and Notation}
\label{subsec:state_definition}
Let $\mathbf{x}_{R}$, $\mathbf{x}_{v}$, and $\mathbf{x}_{g}$ denote the B-spline control points associated with the $\mathrm{SO}(3)$ orientation $\mathbf{R}(t)$, ego-centric velocity $\mathbf{v}(t)$, and local gravity vector $\mathbf{g}(t)$, respectively. The state of the radar-aided proprioceptive estimator is:
\begingroup
\setlength{\savedtextbaselineskip}{\baselineskip}
\small
\setlength{\baselineskip}{\savedtextbaselineskip}
\setlength{\abovedisplayskip}{3pt}
\setlength{\belowdisplayskip}{3pt}
\setlength{\abovedisplayshortskip}{3pt}
\setlength{\belowdisplayshortskip}{3pt}
\begin{equation}
\label{eq:state}
\mathbf{x} \triangleq
\begin{bmatrix}
\mathbf{x}_{R} &
\mathbf{x}_{v} &
\mathbf{x}_{g} &
\mathbf{x}_{b_v} &
\mathbf{x}_{c} &
\mid &
\mathbf{g}^{\mathtt{G}} &
\bar{\mathbf{R}}^{\mathtt{I}_0}_{\mathtt{G}}
\end{bmatrix}
\end{equation}
\vspace{-5mm}
\begin{align*}
\qquad \text{with }\mathbf{x}_{R}
&\triangleq
\begin{bmatrix}
\mathbf{R}^{\mathtt{G}}_{\mathtt{I}_0} &
\cdots &
\mathbf{R}^{\mathtt{G}}_{\mathtt{I}_{l-1}}
\end{bmatrix},
\\
\qquad \mathbf{x}_{v}
&\triangleq
\begin{bmatrix}
\mathbf{v}^{\mathtt{I}_0}_{\mathtt{I}_0} &
\cdots &
\mathbf{v}^{\mathtt{I}_{m-1}}_{\mathtt{I}_{m-1}}
\end{bmatrix},
\\
\qquad \mathbf{x}_{g}
&\triangleq
\begin{bmatrix}
\mathbf{g}^{\mathtt{I}_0} &
\cdots &
\mathbf{g}^{\mathtt{I}_{n-1}}
\end{bmatrix},
\\
\qquad \mathbf{x}_{b_v}
&\triangleq
\begin{bmatrix}
\mathbf{b}_{v_0} &
\cdots &
\mathbf{b}_{v_{i-1}}
\end{bmatrix}
\in \mathbb{R}^{2 \times i},
\\
\qquad \mathbf{x}_{c}
&\triangleq
\begin{bmatrix}
\left\{
\mathbf{R}^{\mathtt{I}}_{\mathtt{R}_u},
\mathbf{p}^{\mathtt{I}}_{\mathtt{R}_u}
\right\}_{u=1}^{N} &
\mathbf{R}^{\mathtt{I}}_{\mathtt{B}} &
\mathbf{p}^{\mathtt{I}}_{\mathtt{B}}
\end{bmatrix},
\\
\qquad \mathbf{g}^{\mathtt{G}}
&\triangleq
\begin{bmatrix}
0 & 0 & 9.81
\end{bmatrix}.
\end{align*}
\endgroup
Here, $\mathbf{R}^{\mathtt{G}}_{\mathtt{I}}$ and $\mathbf{v}^{\mathtt{I}}_{\mathtt{I}}$ denote the IMU orientation 
in the global frame and IMU velocity expressed in the IMU frame, respectively, while $\mathbf{b}_v$ denotes the velocity bias. 
The angular velocity $\boldsymbol{\omega}(t)$ is not included in the state and is obtained from the time derivative of $\mathbf{R}(t)$. 
In our setup, the gyroscope bias \(\mathbf{b}_g\) is only weakly observable and is therefore neglected. Since $\mathbf{v}(t)$ is estimated solely from radar and leg kinematics without an \ac{IMU} acceleration factor, 
the accelerometer bias $\mathbf{b}_a$ is not constrained by 
model and is structurally unobservable;
we therefore set $\mathbf{b}_{a}\!\equiv\!\mathbf{0}$.
The calibration state $\mathbf{x}_{c}$ contains the radar-to-IMU extrinsics $(\mathbf{R}^{\mathtt{I}}_{\mathtt{R}_u}, \mathbf{p}^{\mathtt{I}}_{\mathtt{R}_u})$ for each radar $u \in \{1,\ldots,N\}$ and the body-to-IMU extrinsics $(\mathbf{R}^{\mathtt{I}}_{\mathtt{B}}, \mathbf{p}^{\mathtt{I}}_{\mathtt{B}})$.
Finally, $\mathbf{g}^{\mathtt{G}}$ and $\bar{\mathbf{R}}^{\mathtt{I}_0}_{\mathtt{G}}$ are used to define the gravity-aligned global frame.

\subsection{Slip \& Rolling-Contact-Aware Leg Velocity Factor}
\label{subsec:leg_velocity_factor}

Joint positions and velocities, efforts, contact states, and contact normals are obtained from the robot driver.
Let $\mathcal{L}=\{\mathrm{LF},\mathrm{RF},\mathrm{LH},\mathrm{RH}\}$ be the set of legs, and let $\mathcal{S}_i\subseteq\mathcal{L}$ be the stance legs at time $t_i$. For each stance leg $s\in\mathcal{S}_i$, given the joint angle vector $\tilde{\boldsymbol{\alpha}}_{i,s}$, the nominal foot-center position in the body frame, \(\mathbf{p}^{\mathtt{B}}_{\mathtt{F}_s}\), is computed by forward kinematics:
\begingroup
\setlength{\savedtextbaselineskip}{\baselineskip}
\small
\setlength{\baselineskip}{\savedtextbaselineskip}
\setlength{\abovedisplayskip}{1pt}
\setlength{\belowdisplayskip}{1pt}
\setlength{\abovedisplayshortskip}{1pt}
\setlength{\belowdisplayshortskip}{1pt}
\begin{equation}
\mathbf{p}^{\mathtt{B}_i}_{\mathtt{F}_s}
=
\mathbf{f}_{p,s}\!\left(\tilde{\boldsymbol{\alpha}}_{i,s}\right).
\end{equation}
\endgroup

 
Unlike previous rolling-contact leg odometry~\cite{jung2024co}, which accounts for rounded-foot rolling through the temporal motion of a contact frame before preintegration, we directly impose the kinematic constraint at the terrain-normal support point.
Given the terrain contact normal $\mathbf{n}_{i,s}^{\mathtt{B}}$ and foot radius $r_s$, the support point $\mathbf{p}^{\mathtt{B}_i}_{\mathtt{C}_s}$ and contact-point Jacobian $\mathbf{J}_{c,s}$ are
\begingroup
\setlength{\savedtextbaselineskip}{\baselineskip}
\small
\setlength{\baselineskip}{\savedtextbaselineskip}
\setlength{\abovedisplayskip}{3pt}
\setlength{\belowdisplayskip}{3pt}
\setlength{\abovedisplayshortskip}{3pt}
\setlength{\belowdisplayshortskip}{3pt}
\begin{equation}
\mathbf{p}^{\mathtt{B}_i}_{\mathtt{C}_s}
=
\mathbf{p}^{\mathtt{B}_i}_{\mathtt{F}_s}
-
r_s\mathbf{n}_{i,s}^{\mathtt{B}},
\qquad
\mathbf{J}_{c,s}
=
\mathbf{J}_{p,s}
+
\left\lfloor r_s\mathbf{n}_{i,s}^{\mathtt{B}}\right\rfloor_{\times}
\mathbf{J}_{\omega,s},
\end{equation}
\endgroup
where $\mathbf{J}_{p,s}$ and $\mathbf{J}_{\omega,s}$ are the translational and angular foot Jacobians. This directly accounts for the contact velocity induced by rolling motion.
For each stance leg, the body velocity implied by the contact constraint is
\begingroup
\setlength{\savedtextbaselineskip}{\baselineskip}
\small
\setlength{\baselineskip}{\savedtextbaselineskip}
\setlength{\abovedisplayskip}{3pt}
\setlength{\belowdisplayskip}{3pt}
\setlength{\abovedisplayshortskip}{3pt}
\setlength{\belowdisplayshortskip}{3pt}
\begin{equation}
\label{eq:rolling_leg_velocity}
\mathbf{v}_{\mathtt{B}_i,s}^{\mathtt{B}_i}
=
-\mathbf{J}_{c,s}
\tilde{\dot{\boldsymbol{\alpha}}}_{i,s}
+
\left\lfloor
\mathbf{p}^{\mathtt{B}_i}_{\mathtt{C}_s}
\right\rfloor_{\times}
\boldsymbol{\omega}^{\mathtt{B}}(t_i).
\end{equation}
\endgroup
Thus, unlike finite-difference foot odometry~\cite{noh2026garlileo}, the measurement directly uses the analytical joint-velocity relation.
The angular velocity term is retained as an optimization-dependent quantity, instead of injecting raw gyroscope measurements~\cite{nistico2025muse} or neglecting rotational coupling~\cite{wisth2022vilens}. This provides time-aligned, trajectory-consistent angular velocity without directly injecting raw gyroscope noise.

We encode leg-wise reliability by inflating the encoder-propagated covariance $\mathbf{\Sigma}_{0,i,s}$ using the stance duration $\tau_{i,s}$, joint-effort discontinuity $\Delta e_{i,s}$, and the inter-leg velocity inconsistency $d_{i,s}$ of leg $s$. Here, $d_{i,s}$ is formulated as
\begingroup
\setlength{\savedtextbaselineskip}{\baselineskip}
\small
\setlength{\baselineskip}{\savedtextbaselineskip}
\setlength{\abovedisplayskip}{3pt}
\setlength{\belowdisplayskip}{3pt}
\setlength{\abovedisplayshortskip}{3pt}
\setlength{\belowdisplayshortskip}{3pt}
\begin{equation}
d_{i,s}
=
\frac{1}{\max(|\mathcal{S}_i|-1,1)}
\sum_{r\in\mathcal{S}_i}
\left\|
\mathbf{v}_{\mathtt{B}_i,s}^{\mathtt{B}_i}
-
\mathbf{v}_{\mathtt{B}_i,r}^{\mathtt{B}_i}
\right\|_2^2 .
\end{equation}
\vspace{-2.5mm}
\begin{equation}
\mathbf{\Sigma}_{i,s}
=
\mathbf{\Sigma}_{0,i,s}
+
\left(
\frac{w_{\tau}}{\tau_{i,s}+\epsilon}
+
w_e\Delta e_{i,s}
+
w_d d_{i,s}
\right)\mathbf{I}_3
\end{equation}
\endgroup
downweights the legs after touchdown, abrupt changes in contact, or inconsistent velocity estimates, where $w_{\tau}$, $w_e$, and $w_d$ are constant weights empirically determined.

Let $\mathbf{\Omega}_{i,s}=\mathbf{\Sigma}_{i,s}^{-1}$. To keep the angular coupling dependent on the optimized trajectory, we fuse the joint-velocity term and angular-velocity coupling separately:
\begingroup
\setlength{\savedtextbaselineskip}{\baselineskip}
\small
\setlength{\baselineskip}{\savedtextbaselineskip}
\setlength{\abovedisplayskip}{3pt}
\setlength{\belowdisplayskip}{3pt}
\setlength{\abovedisplayshortskip}{3pt}
\setlength{\belowdisplayshortskip}{3pt}
\begin{equation}
\begin{gathered}
\tilde{\mathbf{v}}_{\mathrm{kin},i}^{\mathtt{B}}
=
\mathbf{\Sigma}_{v_i}
\sum_{s\in\mathcal{S}_i}
\mathbf{\Omega}_{i,s}
\tilde{\mathbf{v}}_{\mathrm{kin},i,s}^{\mathtt{B}},
\\
\text{where}\quad
\mathbf{\Sigma}_{v_i}
=
\left(
\sum_{s\in\mathcal{S}_i}
\mathbf{\Omega}_{i,s}
\right)^{-1},
\quad
\tilde{\mathbf{v}}_{\mathrm{kin},i,s}^{\mathtt{B}}
=
-\mathbf{J}_{c,s}\tilde{\dot{\boldsymbol{\alpha}}}_{i,s}.
\end{gathered}
\end{equation}
\vspace{-2mm}
\begin{equation}
\mathbf{F}_{\omega,i}
=
\mathbf{\Sigma}_{v_i}
\sum_{s\in\mathcal{S}_i}
\mathbf{\Omega}_{i,s}
\mathbf{P}_{i,s},
\quad
\mathbf{P}_{i,s}
=
\left\lfloor
\mathbf{p}^{\mathtt{B}_i}_{\mathtt{C}_s}
\right\rfloor_{\times}.
\end{equation}
\endgroup
The fused leg-derived velocity is 
\(
\mathbf{v}_{\mathtt{B}_i}^{\mathtt{B}_i}
=
\tilde{\mathbf{v}}_{\mathrm{kin},i}^{\mathtt{B}}
+
\mathbf{F}_{\omega,i}\boldsymbol{\omega}^{\mathtt{B}}(t_i)\).
To further handle coherent support-plane slip, we estimate a representative contact normal from the stance legs as follows:
\begingroup
\setlength{\savedtextbaselineskip}{\baselineskip}
\small
\setlength{\baselineskip}{\savedtextbaselineskip}
\setlength{\abovedisplayskip}{3pt}
\setlength{\belowdisplayskip}{3pt}
\setlength{\abovedisplayshortskip}{3pt}
\setlength{\belowdisplayshortskip}{3pt}
\begin{equation}
w_{i,s}
=
\frac{\operatorname{tr}(\mathbf{\Omega}_{i,s})}
{\sum_{r\in\mathcal{S}_i}\operatorname{tr}(\mathbf{\Omega}_{i,r})},
\qquad
\bar{\mathbf{n}}_i
=
\frac{\sum_{s\in\mathcal{S}_i}w_{i,s}\mathbf{n}_{i,s}}
{\left\|\sum_{s\in\mathcal{S}_i}w_{i,s}\mathbf{n}_{i,s}\right\|}.
\end{equation}
\endgroup
An orthonormal tangent basis $\mathbf{T}_i=[\mathbf{t}_{1,i},\mathbf{t}_{2,i}]\in\mathbb{R}^{3\times2}$ is constructed with $\bar{\mathbf{n}}_i^\top\mathbf{T}_i=\mathbf{0}$.
The two-dimensional bias $\mathbf{b}_{v,i}$ is then embedded as $\mathbf{T}_i\mathbf{b}_{v,i}$, allowing shared slip to lie on the local tangent plane. The leg velocity residual ${r}_{L}$ is then
\begingroup
\setlength{\savedtextbaselineskip}{\baselineskip}
\small
\setlength{\baselineskip}{\savedtextbaselineskip}
\setlength{\abovedisplayskip}{3pt}
\setlength{\belowdisplayskip}{3pt}
\setlength{\abovedisplayshortskip}{3pt}
\setlength{\belowdisplayshortskip}{3pt}
\begin{equation}
\mathbf{r}_{L}(t_i)
=
\left(\mathbf{R}^{\mathtt{I}}_{\mathtt{B}}\right)^{\top}
\left(
\mathbf{v}(t_i)
+
\left\lfloor \boldsymbol{\omega}(t_i) \right\rfloor_{\times}
\mathbf{p}^{\mathtt{I}}_{\mathtt{B}}
\right)
-
{\mathbf{v}}_{\mathtt{B}_i}^{\mathtt{B}_i}
-
\mathbf{T}_i\mathbf{b}_{v,i}.
\end{equation}
\endgroup

\subsection{Kinematics-Aware Radar Velocity Factor}
\label{subsec:radar_velocity_factor}
To overcome the asymmetric angular resolution inherent in radar, we employ an orthogonal dual-radar configuration, similar to \cite{huang2024multi}.
The index $u \in \{1, 2\}$ denotes the standard forward-facing and $90^\circ$-rolled radars. For radar $u$ at time $t_i$, $\tilde{\mathbf{p}}^{\mathtt{r}_u}_{i,j}\in\mathbb{R}^3$ and $\tilde{v}^{\mathtt{r}_u}_{m_i,j}\in\mathbb{R}$ denote the $j$-th target location and radial velocity, respectively. The line-of-sight unit direction for each target is defined as $\mathbf{d}_{i,u,j} = \tilde{\mathbf{p}}^{\mathtt{r}_u}_{i,j}/\|\tilde{\mathbf{p}}^{\mathtt{r}_u}_{i,j}\|$.

\begin{figure}[!t]
    \centering
    \includegraphics[trim= 0cm 0.1cm 0cm 0cm, clip,width=1.0\columnwidth]{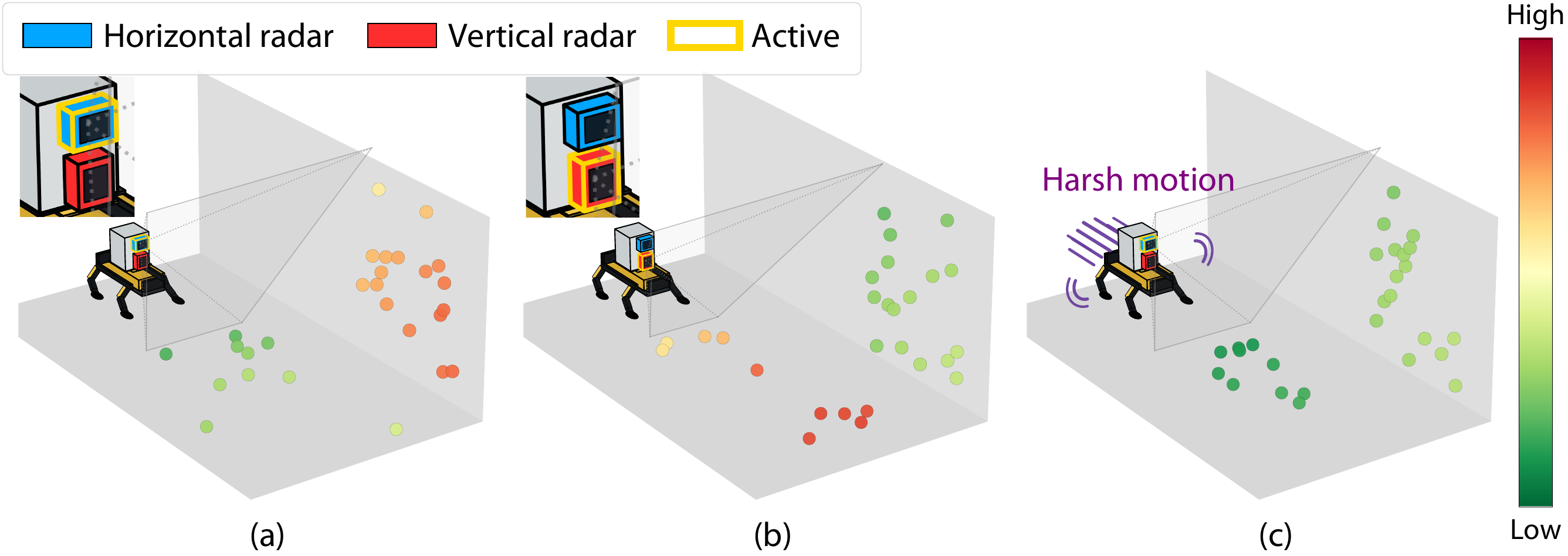}
    \vspace{-5.5mm}
    \caption{Kinematics-aware multi-radar factor. (a) For the horizontal radar, wall points aligned with the azimuth are weighted higher. (b) For the vertical radar, floor and ceiling points aligned with the azimuth are weighted higher. (c) Under harsh motion, increased positional uncertainty lowers the radar Doppler residual weights.}
    \label{fig:radar}
    \vspace{-5.5mm}
\end{figure}

We model the boresight-dependent angular uncertainty as $\mathbf{C}_{\Theta,u,j}=\mathrm{diag}(\sigma_{\theta,u,j}^2,\sigma_{\phi,u,j}^2)$. Propagating it through the unit-vector Jacobian $\mathbf{J}_{\mathbf{d}}$ yields the IMU-frame directional covariance $\mathbf{\Sigma}_{\mathbf{d},u,j}^{\mathtt{I}}=\mathbf{R}^{\mathtt{I}}_{\mathtt{R}_u}\mathbf{J}_{\mathbf{d}}\mathbf{C}_{\Theta,u,j}\mathbf{J}_{\mathbf{d}}^\top(\mathbf{R}^{\mathtt{I}}_{\mathtt{R}_u})^\top$. 
A key novelty of our formulation is the kinematics-aware orthogonal covariance, which accounts for the way directional uncertainty affects the 1D radial Doppler measurement given the robot's instantaneous motion, as shown in \figref{fig:radar}. We project the 3D directional covariance onto the scalar Doppler axis using ego-velocity, yielding the state-dependent Doppler variance
\begingroup
\setlength{\savedtextbaselineskip}{\baselineskip}
\small
\setlength{\baselineskip}{\savedtextbaselineskip}
\setlength{\abovedisplayskip}{3pt}
\setlength{\belowdisplayskip}{3pt}
\setlength{\abovedisplayshortskip}{3pt}
\setlength{\belowdisplayshortskip}{3pt}
\begin{equation}
\label{eq:radar_dynamic_variance}
\sigma_{d, u, j}^2(t_i) = \sigma_{v_d}^2 + \mathbf{v}_{\mathtt{R}_u}(t_i)^{\top} \mathbf{\Sigma}_{\mathbf{d}, u, j}^{\mathtt{I}} \mathbf{v}_{\mathtt{R}_u}(t_i),
\end{equation}
\endgroup
where $\sigma_{v_d}^2$ is the baseline Doppler noise, and $\mathbf{v}_{\mathtt{R}_u}(t_i) = \mathbf{v}(t_i) + \lfloor \boldsymbol{\omega}(t_i) \rfloor_{\times} \mathbf{p}^{\mathtt{I}}_{\mathtt{R}_u}$.
The radar velocity residual 
is given by
\begingroup
\setlength{\savedtextbaselineskip}{\baselineskip}
\small
\setlength{\baselineskip}{\savedtextbaselineskip}
\setlength{\abovedisplayskip}{3pt}
\setlength{\belowdisplayskip}{3pt}
\setlength{\abovedisplayshortskip}{3pt}
\setlength{\belowdisplayshortskip}{3pt}
\begin{equation}
\label{eq:radar_velocity_residual_scalar}
r_{R,u,j}(t_i)
=
-
\mathbf{d}_{i,u,j}^{\top}
\left(\mathbf{R}^{\mathtt{I}}_{\mathtt{R}_u}\right)^{\top}
\left(
\mathbf{v}(t_i)
+
\lfloor \boldsymbol{\omega}(t_i) \rfloor_{\times}
\mathbf{p}^{\mathtt{I}}_{\mathtt{R}_u}
\right)
-
\tilde{v}^{\mathtt{r}_u}_{m_i,j}.
\end{equation}
\endgroup
Stacking all targets across both radars yields the overall kinematics-aware multi-radar velocity residual $\mathbf{r}_{R}(t_i)$. 

\subsection{Factor Graph Optimization}
\label{subsec:factor_graph_optimization}
Let $W_i$ denote the active window at iteration $i$. 
The radar-aided proprioceptive spline optimization is formulated as
\begingroup
\setlength{\savedtextbaselineskip}{\baselineskip}
\small
\setlength{\baselineskip}{\savedtextbaselineskip}
\setlength{\abovedisplayskip}{3pt}
\setlength{\belowdisplayskip}{3pt}
\setlength{\abovedisplayshortskip}{3pt}
\setlength{\belowdisplayshortskip}{3pt}
\begin{align}
\min_{\mathbf{x}}
\sum_{k\in W_i}
\Bigl\{
& \left\|\mathbf{r}_{L}(t_k)\right\|_{\mathbf{\Omega}_{L,k}}^{2}
+
\rho_{r}\!\left(
\left\|\mathbf{r}_{R}(t_k)\right\|_{\mathbf{\Omega}_{R,k}}^{2}
\right)
\\
& +
\left\|\mathbf{r}_{G}(t_k)\right\|_{\mathbf{\Omega}_{G}}^{2}
+
\left\|\mathbf{r}_{\mathcal{S}^{2}}(t_k)\right\|_{\mathbf{\Omega}_{\mathcal{S}^{2}}}^{2}
+
\left\|\mathbf{r}_{\omega}(t_k)\right\|_{\mathbf{\Omega}_{\omega}}^{2}
 \Bigr\}
\nonumber \\
&  +
\left\|\mathbf{r}_{b}\right\|_{\mathbf{\Omega}_{b}}^{2}
+
\left\|\mathbf{r}_{e}\right\|_{\mathbf{\Omega}_{e}}^{2}
+
\left\|\mathbf{r}_{m}\right\|_{\mathbf{\Omega}_{m}}^{2}
+
\mathds{1}_{\mathrm{calib}}
\left\|\mathbf{r}_{C}\right\|_{\mathbf{\Omega}_{C}}^{2},\nonumber
\end{align}
\endgroup
where $\|\mathbf{r}\|_{\mathbf{\Omega}}^{2} \triangleq \mathbf{r}^{\top}\mathbf{\Omega}\mathbf{r}$, and $\rho_r(\cdot)$ denotes the Cauchy loss, which suppresses dynamic radar outliers and enables stable velocity estimation with $\mathbf{r}_{L}$ without costly RANSAC.
$\mathds{1}_{\mathrm{calib}}$ enables online multi-sensor extrinsic calibration.
The corresponding calibration prior residual is
\begingroup
\setlength{\savedtextbaselineskip}{\baselineskip}
\small
\setlength{\baselineskip}{\savedtextbaselineskip}
\setlength{\abovedisplayskip}{3pt}
\setlength{\belowdisplayskip}{3pt}
\setlength{\abovedisplayshortskip}{3pt}
\setlength{\belowdisplayshortskip}{3pt}
\begin{equation}
\mathbf{r}_{C}
=
\begin{bmatrix}
\left\{
\begin{bmatrix}
\mathbf{r}_{\mathrm{rot}}(
\mathbf{R}^{\mathtt{I}}_{\mathtt{R}_u},
\hat{\mathbf{R}}^{\mathtt{I}}_{\mathtt{R}_u})^{\top} &
\mathbf{r}_{\mathrm{pos}}(
\mathbf{p}^{\mathtt{I}}_{\mathtt{R}_u},
\hat{\mathbf{p}}^{\mathtt{I}}_{\mathtt{R}_u})^{\top}
\end{bmatrix}^{\top}
\right\}_{u=1}^{N}
\\[1mm]
\begin{bmatrix}
\mathbf{r}_{\mathrm{rot}}(
\mathbf{R}^{\mathtt{I}}_{\mathtt{B}},
\hat{\mathbf{R}}^{\mathtt{I}}_{\mathtt{B}})^{\top} &
\mathbf{r}_{\mathrm{pos}}(
\mathbf{p}^{\mathtt{I}}_{\mathtt{B}},
\hat{\mathbf{p}}^{\mathtt{I}}_{\mathtt{B}})^{\top}
\end{bmatrix}^{\top}
\end{bmatrix},
\end{equation}
\endgroup
with
\(
\mathbf{r}_{\mathrm{rot}}(\mathbf{R},\hat{\mathbf{R}})
=
\operatorname{Log}(\hat{\mathbf{R}}^{-1}\mathbf{R})^{\vee}
\)
and
\(
\mathbf{r}_{\mathrm{pos}}(\mathbf{p},\hat{\mathbf{p}})
=
\mathbf{p}-\hat{\mathbf{p}}
\).
We set $\mathbf{\Omega}_{L,k}=\mathbf{\Sigma}_{v_k}^{-1}$ from the fused leg-velocity covariance to downweight unreliable stances.
The radar information matrix $\mathbf{\Omega}_{R,k}$ is a diagonal matrix with kinematics-aware weights $\Omega_{R,u,j}(t_k) = \sigma_{d,u,j}^{-2}(t_k)$ from \eqref{eq:radar_dynamic_variance}.
The residuals $\mathbf{r}_{G}(\cdot)$, $\mathbf{r}_{\mathcal{S}^{2}}(\cdot)$, $\mathbf{r}_{\omega}(\cdot)$, $\mathbf{r}_{b}$, $\mathbf{r}_{e}$, and $\mathbf{r}_{m}$ follow GaRLILEO~\cite{noh2026garlileo},
whereas $\mathbf{r}_{L}(\cdot)$, $\mathbf{r}_{R}(\cdot)$, and $\mathbf{r}_{C}$ are newly introduced.
\section{Exteroceptive-based State Estimator}
\subsection{State Definition and Notation}
The state in the exteroceptive-based estimator is
\begingroup
\setlength{\savedtextbaselineskip}{\baselineskip}
\small
\setlength{\baselineskip}{\savedtextbaselineskip}
\setlength{\abovedisplayskip}{3pt}
\setlength{\belowdisplayskip}{3pt}
\setlength{\abovedisplayshortskip}{3pt}
\setlength{\belowdisplayshortskip}{3pt}
\begin{equation}
\label{eq:state_vector}
\mathbf{x}_k
\triangleq
\begin{bmatrix}
\mathbf{p}^{\mathtt{G}}_{\mathtt{I},k} &
\mathbf{R}^{\mathtt{G}}_{\mathtt{I},k} &
\mathbf{v}^{\mathtt{G}}_{\mathtt{I},k} &
\mathbf{b}_{g,k} &
\mathbf{b}_{a,k}
\end{bmatrix}.
\end{equation}
\endgroup
Here, \(\mathbf{p}^{\mathtt{G}}_{\mathtt{I},k}\), \(\mathbf{R}^{\mathtt{G}}_{\mathtt{I},k}\), and \(\mathbf{v}^{\mathtt{G}}_{\mathtt{I},k}\) denote the IMU position, orientation, and velocity in the global frame, respectively, while \(\mathbf{b}_{g,k}\) and \(\mathbf{b}_{a,k}\) denote the gyroscope and accelerometer biases. Unlike in Section~\ref{subsec:state_definition}, \(\mathbf{b}_{a,k}\) is now observable and modeled as a random walk, whereas \(\mathbf{b}_{g,k}\) remains neglected. The state is estimated at each camera timestamp \(k\).
\subsection{Spatial-Feature-Aware Image Enhancement}
\label{sec:image_enhancement}
AGCWD~\cite{huang2012efficient} and CLAHE~\cite{zuiderveld1994contrast} are widely used for contrast enhancement under non-uniform illumination.
Unlike fixed-parameter weighted fusion,
we optimize their parameters online via \ac{BO}.
Let \(I_t^c\) be the intensity image from camera \(c\) at time \(t\), and let \(\theta=[\eta,\kappa]^\top\) denote the enhancement parameters, where \(\eta\) and \(\kappa\) are the AGCWD weighting parameter and CLAHE clip limit, respectively. 
The enhanced images are \(I_{\mathrm{A}}^c = \mathcal{A}(I_t^c;\eta)\) and \(I_{\mathrm{C}}^c = \mathcal{C}(I_t^c;\kappa,g)\), where \(\mathcal{A}(\cdot)\) and \(\mathcal{C}(\cdot)\) denote AGCWD and CLAHE, with fixed CLAHE tile-grid size \(g\).

Let $\mu(\cdot)$ and $\mathcal{H}(\cdot)$ denote the image mean and histogram entropy, respectively. The enhanced outputs are fused as
\begingroup
\setlength{\savedtextbaselineskip}{\baselineskip}
\small
\setlength{\baselineskip}{\savedtextbaselineskip}
\setlength{\abovedisplayskip}{3pt}
\setlength{\belowdisplayskip}{3pt}
\setlength{\abovedisplayshortskip}{3pt}
\setlength{\belowdisplayshortskip}{3pt}
\begin{equation}
\label{eq:image_fusion}
\begin{gathered}
\mathcal{E}(I_t^c;\theta)
=
\rho I_{\mathrm{A}}^c + (1-\rho) I_{\mathrm{C}}^c, \\
\rho
=
\frac{\mu(I_{\mathrm{A}}^c)}
{2\left(\mu(I_{\mathrm{A}}^c)+\mu(I_{\mathrm{C}}^c)+\epsilon\right)}
+
\frac{\mathcal{H}(I_{\mathrm{A}}^c)}
{2\left(\mathcal{H}(I_{\mathrm{A}}^c)+\mathcal{H}(I_{\mathrm{C}}^c)+\epsilon\right)} .
\end{gathered}
\end{equation}
\endgroup

We optimize the enhancement parameters via \ac{BO} using the proposed Spatial-Feature-Aware Entropy-Weighted Gradient (SFA-EWG) utility. Unlike EWG~\cite{kim2020proactive}, whose penalty is scaled by the unweighted image-gradient mean and can become negative, we use a positive EWG score \(U_{\mathrm{P\text{-}EWG}}\), whose penalty is proportional to the entropy-weighted gradient:
\begingroup
\setlength{\savedtextbaselineskip}{\baselineskip}
\small
\setlength{\baselineskip}{\savedtextbaselineskip}
\setlength{\abovedisplayskip}{3pt}
\setlength{\belowdisplayskip}{3pt}
\setlength{\abovedisplayshortskip}{3pt}
\setlength{\belowdisplayshortskip}{3pt}
\begin{equation}
\begin{gathered}
G_W=\sum_{i=1}^{N} W_i g_i,
\qquad
\pi(H_i)=\frac{2}{1+\exp(-\alpha H_i+\tau)}-1,\\
M_i(H_i)=
\begin{cases}
1, & 0\le H_i < H_{\mathrm{th}},\\
0, & H_{\mathrm{th}}\le H_i \le 1,
\end{cases}\\
U_{\mathrm{P\text{-}EWG}}
=
\sum_{i=1}^{N}
\left[
W_i g_i
+
\frac{G_W}{N}\pi(H_i)M_i(H_i)
\right].
\end{gathered}
\end{equation}
\endgroup
Here, \(g_i\) and \(H_i\) denote the image gradient and entropy of pixel \(i\), respectively, \(\pi(\cdot)\) is an activation function, and \(M_i(\cdot)\) is a binary saturation mask used to penalize saturated regions. Since \(G_W\ge0\) and 
 \(\pi(H_i)M_i(H_i)\ge-1\), this formulation ensures \(U_{\mathrm{P\text{-}EWG}}\ge0\). SFA-EWG further incorporates BRISK~\cite{leutenegger2011brisk} feature count and spatial uniformity, aligning the objective with feature-based visual SLAM:
\begingroup
\setlength{\savedtextbaselineskip}{\baselineskip}
\small
\setlength{\baselineskip}{\savedtextbaselineskip}
\setlength{\abovedisplayskip}{3pt}
\setlength{\belowdisplayskip}{3pt}
\setlength{\abovedisplayshortskip}{3pt}
\setlength{\belowdisplayshortskip}{3pt}
\begin{equation}
J_t(\theta)
=
\frac{1}{|\mathcal{C}_t|}
\sum_{c\in\mathcal{C}_t}
U_{\mathrm{SFA\text{-}EWG}}
\left(
\mathcal{E}(I_t^c;\theta)
\right),
\end{equation}
\endgroup
where \(\mathcal{C}_t\) is the set of available stereo cameras. We define
\(U_{\mathrm{SFA\text{-}EWG}}
=
U_{\mathrm{P\text{-}EWG}}
\left(1+\lambda_f Q_f Q_s\right),\)
where \(Q_f\) and \(Q_s\) denote feature-count quality and feature spatial uniformity, respectively. 
The optimal parameters are obtained by maximizing the objective, while the next sample is selected by expected improvement from the updated Gaussian process:
\begingroup
\setlength{\savedtextbaselineskip}{\baselineskip}
\small
\setlength{\baselineskip}{\savedtextbaselineskip}
\setlength{\abovedisplayskip}{3pt}
\setlength{\belowdisplayskip}{3pt}
\setlength{\abovedisplayshortskip}{3pt}
\setlength{\belowdisplayshortskip}{3pt}
\begin{equation}
\theta_t^\star = \arg\max_{\theta} J_t(\theta),
\qquad
\theta_{n+1} = \arg\max_{\theta} \mathrm{EI}(\theta).
\end{equation}
\endgroup

\begin{figure}[!t]
    \centering
    \includegraphics[trim= 0.1cm 0.1cm 0.1cm 0.1cm, clip,width=1.0\columnwidth]{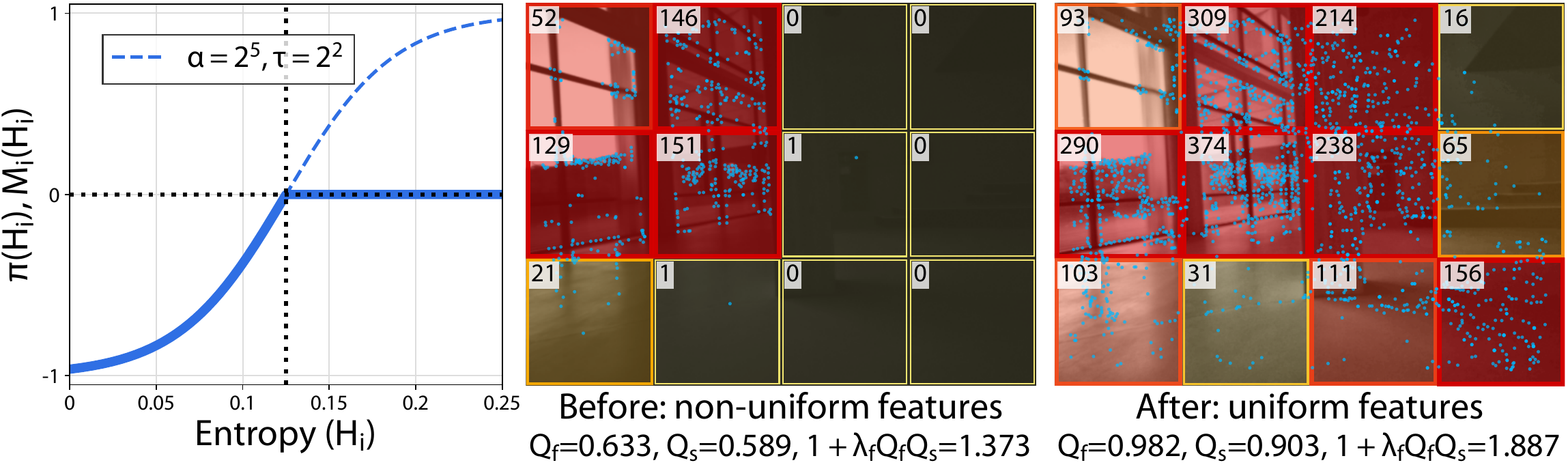}
    \vspace{-5.5mm}
    \caption{\textbf{Left}: Plot of $\pi(H_i)M_i(H_i)$. \textbf{Middle}: Before enhancement, features are non-uniformly distributed.
    \textbf{Right}: After enhancement, more features are extracted and more uniformly distributed.}
    \label{fig:image}
    \vspace{-5.5mm}
\end{figure}

To reduce computation cost, BO is triggered only when the raw-image utility changes sufficiently:
\begingroup
\setlength{\savedtextbaselineskip}{\baselineskip}
\small
\setlength{\baselineskip}{\savedtextbaselineskip}
\setlength{\abovedisplayskip}{3pt}
\setlength{\belowdisplayskip}{3pt}
\setlength{\abovedisplayshortskip}{3pt}
\setlength{\belowdisplayshortskip}{3pt}
\begin{equation}
d_t=
\frac{
|U_{\mathrm{raw},t}-U_{\mathrm{raw},t-1}|
}{
|U_{\mathrm{raw},t}|+|U_{\mathrm{raw},t-1}|+\epsilon
}.
\end{equation}
\endgroup
If \(d_t\) exceeds a threshold, BO updates \(\theta_t^\star\); otherwise, the previous parameters are reused. 
As shown in \figref{fig:image}, the enhanced image exhibits dense, spatially uniform features.

\begin{figure}[!t]
    \centering
    \includegraphics[trim= 5.5cm 2.7cm 5.5cm 3.0cm, clip,width=0.9\columnwidth]{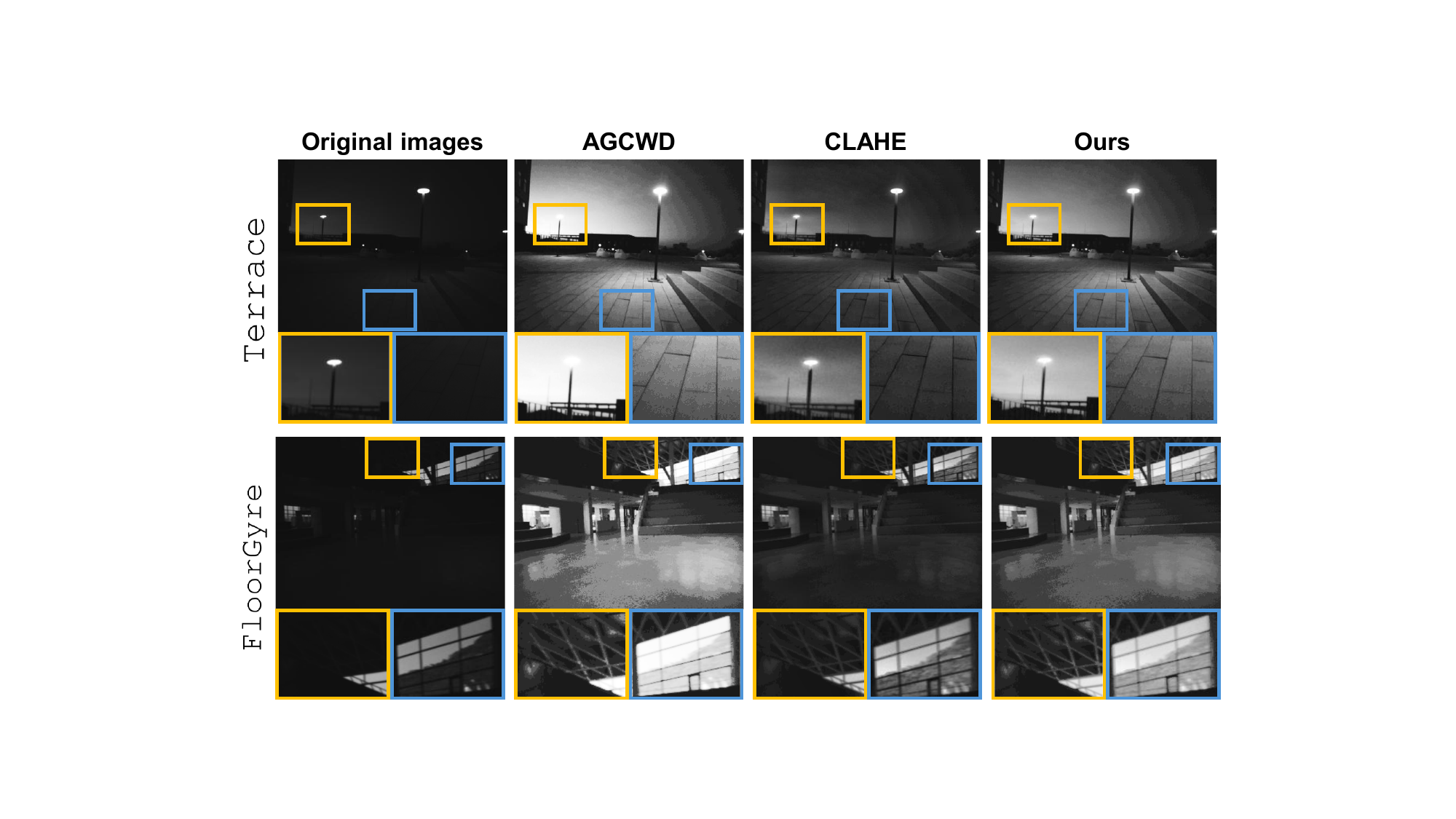}
    \vspace{-2mm}
    \caption{Qualitative comparison of image enhancement results.}
    \label{fig:enhancement}
    \vspace{-5mm}
\end{figure}

As shown in \figref{fig:enhancement}, AGCWD tends to over-enhance bright regions, whereas CLAHE tends to leave dark regions under-enhanced.
In contrast, our method optimizes the parameters using \ac{BO}, encouraging features to be densely and uniformly distributed across the image. As a result, it provides stable enhancement without over- or under-correction.

\subsection{Visual Frontend}
\subsubsection{Zero-Velocity Detection}
Near-static intervals are detected from proprioceptive estimates, during which keyframe insertion is suppressed. This reduces redundant frame processing and maintains a well-distributed sliding window.

\subsubsection{Gravity-Enhanced Keypoint Extraction}
The visual frontend extracts BRISK~\cite{leutenegger2011brisk} keypoints and descriptors from each image. Unlike the original BRISK formulation, we orient the descriptors using our estimated gravity direction, obtained from proprioceptive-based local gravity estimation.
\subsubsection{Matching and Initialization}
New 3D landmarks are initialized via stereo triangulation across keyframe images, as well as through triangulation between the current live frame and previously selected keyframes.
We initialize the relative pose through dead reckoning using the velocity and rotation estimates from the proprioceptive estimator.

\subsubsection{Motion-Inconsistent Landmark Rejection}
For each candidate match, the current relative camera pose, constrained by the proprioceptive motion factors, defines the expected epipolar geometry. 
Matches violating epipolar consistency are discarded without landmark creation.

\subsection{Visual-Inertial SLAM}
The exteroceptive state estimator builds on the OKVIS2-X backend~\cite{boche2025okvis2}. 
We add proprioceptive relative-pose and velocity factors to inject proprioceptive motion constraints.

\subsubsection{Proprioceptive Relative-Pose Factor}
For two camera states \(k-1\) and \(k\), we use the proprioceptive spline to compute the time-aligned relative transform between the corresponding camera timestamps. This relative transform,
\({}^{P}\mathbf{T}^{\mathtt{I},k-1}_{\mathtt{I},k}\),
defines the proprioceptive relative-pose factor:
\begingroup
\setlength{\savedtextbaselineskip}{\baselineskip}
\small
\setlength{\baselineskip}{\savedtextbaselineskip}
\setlength{\abovedisplayskip}{3pt}
\setlength{\belowdisplayskip}{3pt}
\setlength{\abovedisplayshortskip}{3pt}
\setlength{\belowdisplayshortskip}{3pt}
\begin{equation}
\mathbf{e}^{k}_{P,r}
=
\operatorname{Log}\!\left(
\left({}^{P}\mathbf{T}^{\mathtt{I},k-1}_{\mathtt{I},k}\right)^{-1}
\left[
\left(\mathbf{T}^{\mathtt{G}}_{\mathtt{I},k-1}\right)^{-1}
\mathbf{T}^{\mathtt{G}}_{\mathtt{I},k}
\right]
\right)^{\vee},
\end{equation}
\endgroup
where \(\operatorname{Log}(\cdot)\) is the logarithm map from \(\mathrm{SE}(3)\) to \(\mathfrak{se}(3)\).

\subsubsection{Velocity Factor}
The proprioceptive estimator provides the ego-centric velocity  
\({}^{P}{\mathbf{v}}^{\mathtt{I}}_{\mathtt{I},k}\), which is compared against the velocity state of the exteroceptive estimator as
\begingroup
\setlength{\savedtextbaselineskip}{\baselineskip}
\small
\setlength{\baselineskip}{\savedtextbaselineskip}
\setlength{\abovedisplayskip}{3pt}
\setlength{\belowdisplayskip}{3pt}
\setlength{\abovedisplayshortskip}{3pt}
\setlength{\belowdisplayshortskip}{3pt}
\begin{equation}
\mathbf{e}^{k}_{P,v}
=
{}^{P}{\mathbf{v}}^{\mathtt{I}}_{\mathtt{I},k}
-
\left(\mathbf{R}^{\mathtt{G}}_{\mathtt{I},k}\right)^{\top}
\mathbf{v}^{\mathtt{G}}_{\mathtt{I},k}.
\end{equation}
\endgroup

\subsubsection{Loop Closure}
RAGNAROK maintains a DBoW2~\cite{galvez2012bags} database. Inspired by SHeRLoc~\cite{kim2025sherloc}, each radar scan is converted into a polar BEV image, and an FFT-based cross-correlation score is used with the DBoW2 score to rank candidates under visual degradation. A candidate is accepted only after geometric verification with 3D-2D RANSAC. Once verified, the corresponding pose-graph edges are recovered as landmarks and observations, and new observations are added between the current and matched frames.
\subsection{Factor Graph Optimization}
All factors are jointly estimated by minimizing the following nonlinear least-squares objective:
\begingroup
\setlength{\savedtextbaselineskip}{\baselineskip}
\small
\setlength{\baselineskip}{\savedtextbaselineskip}
\setlength{\abovedisplayskip}{3pt}
\setlength{\belowdisplayskip}{3pt}
\setlength{\abovedisplayshortskip}{3pt}
\setlength{\belowdisplayshortskip}{3pt}
\begin{align}
c(\mathbf{x}) =
&\frac{1}{2}
\sum_{i}\sum_{k\in\mathcal{K}}\sum_{j\in\mathcal{J}(i,k)}
\rho_c\!\left(
\mathbf{e}^{i,j,k}_r{}^{\!\top}
\mathbf{W}_r
\mathbf{e}^{i,j,k}_r
\right)
\nonumber\\
&+
\frac{1}{2}
\sum_{k\in\mathcal{I}}
\mathbf{e}^{k}_{I}{}^{\!\top}
\mathbf{W}^{k}_{I}
\mathbf{e}^{k}_{I}
+
\frac{1}{2}
\sum_{(a,b)\in\mathcal{P}}
\mathbf{e}^{a,b}_{p}{}^{\!\top}
\mathbf{W}^{a,b}_{p}
\mathbf{e}^{a,b}_{p}
\nonumber\\
&+
\frac{1}{2}
\sum_{m\in\mathcal{M}}
\rho_t\!\left(
\mathbf{e}_{m}^{\top}\mathbf{W}_{m}\mathbf{e}_{m}
\right)
+
\frac{1}{2}
\sum_{k\in\mathcal{R}_{P}}
\mathbf{e}^{k}_{P,r}{}^{\!\top}
\mathbf{W}^{k}_{P,r}
\mathbf{e}^{k}_{P,r}
\nonumber\\
&+
\frac{1}{2}
\sum_{k\in\mathcal{V}_{P}}
\mathbf{e}^{k}_{P,v}{}^{\!\top}
\mathbf{W}^{k}_{P,v}
\mathbf{e}^{k}_{P,v}.
\label{eq:overall_objective}
\end{align}
\endgroup
Here, \(\mathcal{R}_{P}\) and \(\mathcal{V}_{P}\) denote the sets of states to which valid \(\mathbf{e}^{k}_{P,r}\) and \(\mathbf{e}^{k}_{P,v}\) are added. Following OKVIS2-X~\cite{boche2025okvis2}, \(\mathbf{e}_r\), \(\mathbf{e}_I\), \(\mathbf{e}_p\), and \(\mathbf{e}_m\) denote the visual reprojection, IMU, relative pose-graph, and map-alignment residuals, respectively, while predicted depth maps are fused into local submaps for occupancy mapping. \(\mathbf{e}_I\) is retained for stable estimation until proprioceptive constraints become available. To mitigate potential overconfidence, we decouple the velocity spline from accelerometer measurements, retain the angular-velocity terms in \(\mathbf{r}_{L}\) and \(\mathbf{r}_{R}\) as optimization-dependent quantities rather than directly injecting gyroscope measurements, and downweight the rotational component of \(\mathbf{e}^{k}_{P,r}\). Empirically, retaining \(\mathbf{e}_I\) improves estimation performance.

\subsection{Adaptive Weighting}
In Eq.~\eqref{eq:overall_objective}, $\mathbf{W}_{r}$, $\mathbf{W}^{k}_{I}$, $\mathbf{W}^{a,b}_{p}$, and $\mathbf{W}_{m}$ are computed from the keypoint size, linearized error propagation, marginalization, and occupancy gradient, respectively~\cite{boche2025okvis2}, whereas $\mathbf{W}^{i}_{P,r}$ and $\mathbf{W}^{i}_{P,v}$ are initialized with fixed nominal information matrices reflecting their relative confidence. RAGNAROK then adapts these weights online according to visual reliability. For camera $c$, each landmark covers a circle of radius $r_c=\kappa\min(H_c,W_c)$ and area $a_c=\pi r_c^2$, where $\kappa$ is the coverage-radius coefficient and $H_c,W_c$ are the image dimensions. Then, visual tracking quality $Q_i$ is defined as
\begingroup
\setlength{\savedtextbaselineskip}{\baselineskip}
\small
\setlength{\baselineskip}{\savedtextbaselineskip}
\setlength{\abovedisplayskip}{3pt}
\setlength{\belowdisplayskip}{3pt}
\setlength{\abovedisplayshortskip}{3pt}
\setlength{\belowdisplayshortskip}{3pt}
\begin{equation}
Q_i =
\begin{cases}
0, & M_i < 8,\\[1mm]
\displaystyle
\frac{\sum_c \max(0,A_{c,i}-a_c)}
{\sum_c (H_cW_c-a_c)}, & M_i \geq 8,
\end{cases}
\end{equation}
\endgroup
where $M_i$ is the number of tracked keypoints and $A_{c,i}$ is the union coverage area. We define $\alpha_i=1/\max(Q_i,Q_{\min})$, with $Q_{\min}=0.1$, and scale the proprioceptive information as $\widetilde{\mathbf{W}}^{i}_{P,r}=\alpha_i\mathbf{W}^{i}_{P,r}$ and $\widetilde{\mathbf{W}}^{i}_{P,v}=\alpha_i\mathbf{W}^{i}_{P,v}$. Thus, lower visual quality increases the proprioceptive contribution up to $1/Q_{\min}$. Visual fallback is activated when $Q_i<Q_{\mathrm{th}}$ or $U_{\mathrm{SFA\text{-}EWG},i}<U_{\mathrm{th}}$. Let \(\delta_i\) be 1 during visual fallback and 0 otherwise. Given nominal gains \(\gamma_P,\gamma_E>1\), the proprioceptive matrices \(\mathbf{W}^{i}_{P,r}\) and \(\mathbf{W}^{i}_{P,v}\) are further scaled by \(\gamma_P^{\delta_i}\), whereas the exteroceptive matrices \(\mathbf{W}_{r}\), \(\mathbf{W}_{p}^{a,b}\), and \(\mathbf{W}_{m}\) are scaled by \(\gamma_E^{-\delta_i}\). This prioritizes proprioception and suppresses unreliable visual constraints during fallback.

%% file: 4_experiment.tex
\section{Experiment}
\label{sec:experiment}

\input{tab/dataset}
\subsection{Dataset}
\begin{figure}[!t]
    \centering
    \includegraphics[trim= 1.6cm 1.6cm 1.6cm 2.1cm, clip,width=0.87\columnwidth]{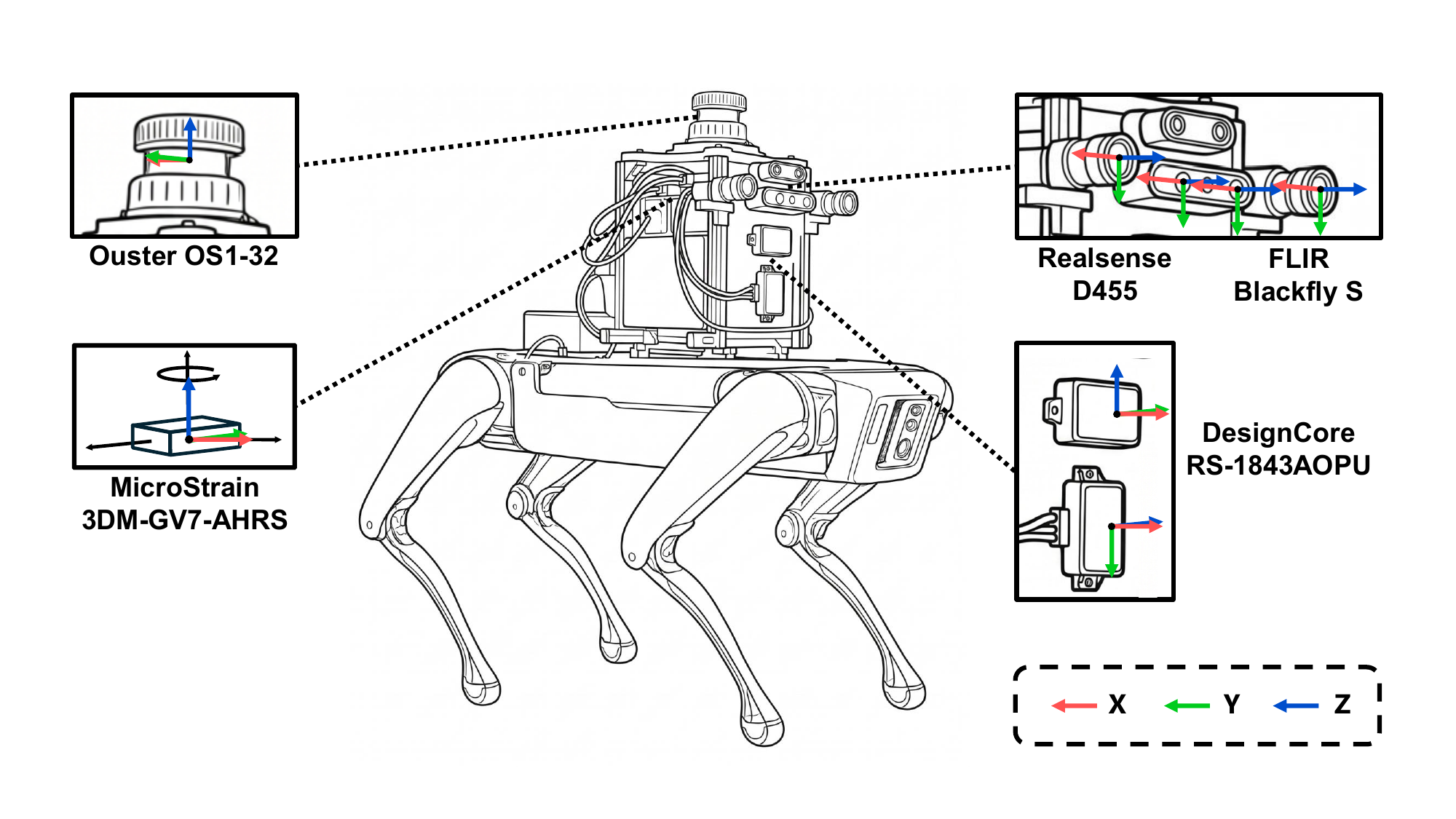}    
    \vspace{-2mm}
    \caption{System overview of the platform and coordinates of sensors.}
    \label{fig:robot}
    \vspace{-3mm}
\end{figure}

RAGNAROK requires multi-radar, stereo camera, \ac{IMU}, and encoder data.
As shown in \tabref{table:dataset}, existing legged-robot datasets lack either radar or camera data. To address this gap, we collected a real-world dataset in challenging environments, as summarized in \tabref{tab:sequence}, using the Boston Dynamics Spot
\figref{fig:robot}.
Ground-truth trajectories were generated using PALoc~\cite{hu2024paloc} with Leica RTC360 TLS maps exhibiting Bundle Errors of $0.002$--$0.006\,\mathrm{m}$. The initial pose was refined using ICP.
We further evaluate the proposed proprioceptive estimator on two public datasets, GaRLILEO~\cite{noh2026garlileo} and Co-RaL~\cite{jung2024co}.

\input{tab/sequences}

\subsection{Experimental Setup}
We evaluate the proposed method through component-level ablations and comparisons with \ac{SOTA} baselines.

\begin{itemize}
    \item \textbf{RAGNAROK-RVI}: single-radar, monocular-visual, inertial configuration, aligned with RVI baselines~\cite{doer2021radar,nissov2026tightly}.
    \item \textbf{RAGNAROK-RKI}: dual-radar, kinematic, inertial configuration without visual measurements.
    \item \textbf{RAGNAROK-VKI}: stereo-visual, kinematic, inertial configuration without radar measurements.
    \item \textbf{RAGNAROK-Odom}: the full odometry configuration with dual-radar, stereo-visual, kinematic, inertial.
    \item \textbf{RAGNAROK-SLAM}: the full SLAM configuration with submapping, loop closure, and bundle adjustment.
\end{itemize}

Odometry accuracy is evaluated using the \ac{RMSE} of \ac{APE} reported for both translational and rotational components. 
The corresponding units are \ac{APE}$_t$ in \meter, and \ac{APE}$_r$ in \degree. 



\subsection{Performance on Self-Collected Dataset}
\begin{figure}[!t]
    \centering
    \includegraphics[trim= 3.1cm 1.5cm 2.9cm 2.2cm, clip,width=0.87\columnwidth]{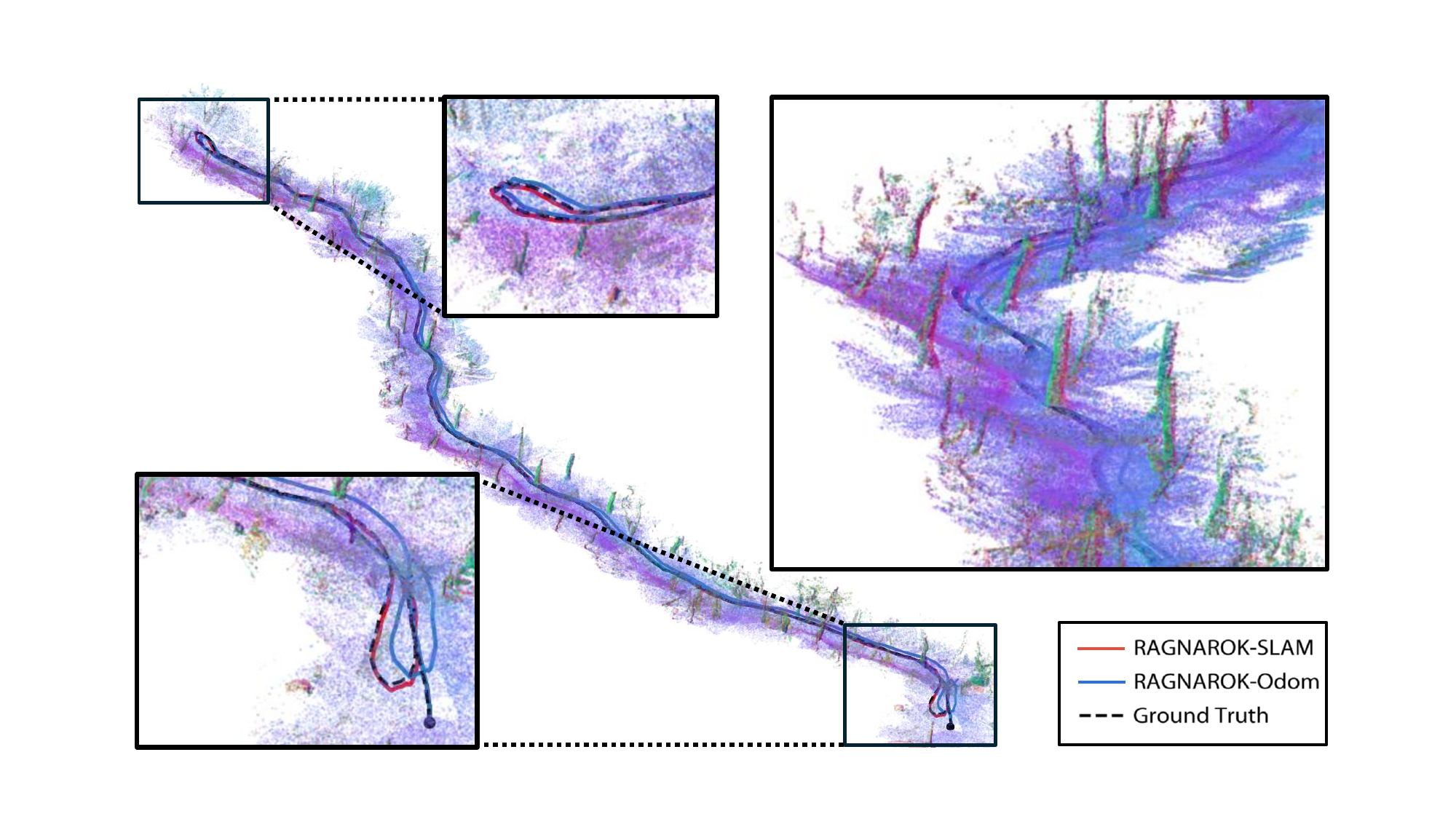}
    \vspace{-2mm}
    \caption{Reconstructed map of \texttt{Mountain-Long} with trajectories.}
    \label{fig:mountain}
    \vspace{-7mm}
\end{figure}

\begin{figure}[!t]
    \centering
    \includegraphics[trim= 0.0cm 1.3cm 0.0cm 2.0cm, clip,width=0.95\columnwidth]{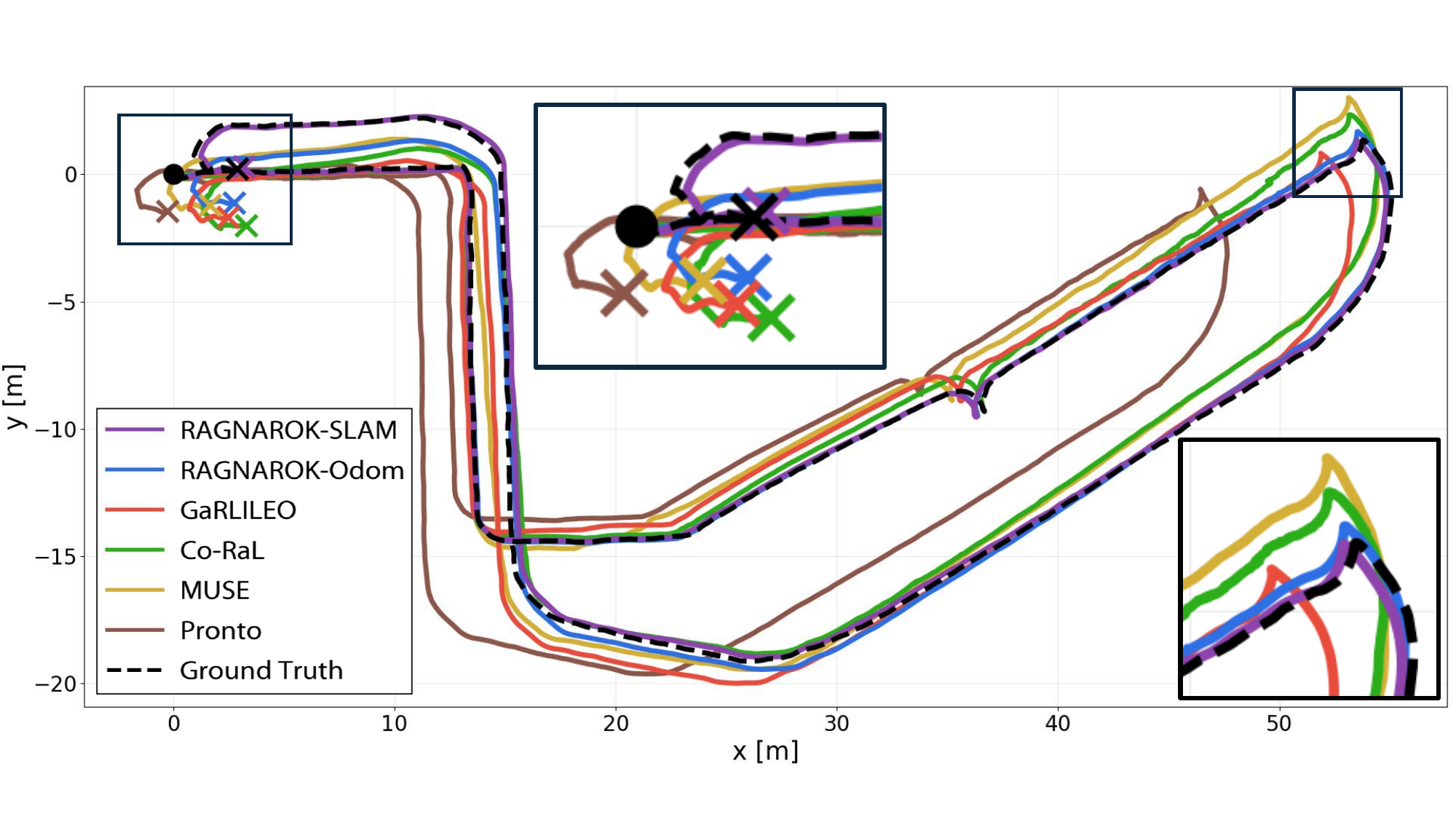}
    \vspace{-2mm}
    \caption{Trajectory comparison with \ac{SOTA} methods on \texttt{Overpass}.}
    \label{fig:overpass}
    \vspace{-3mm}
\end{figure}

\begin{figure}[!t]
    \centering
    \includegraphics[trim= 0cm 5.0cm 0cm 4.9cm, clip,width=0.95\columnwidth]{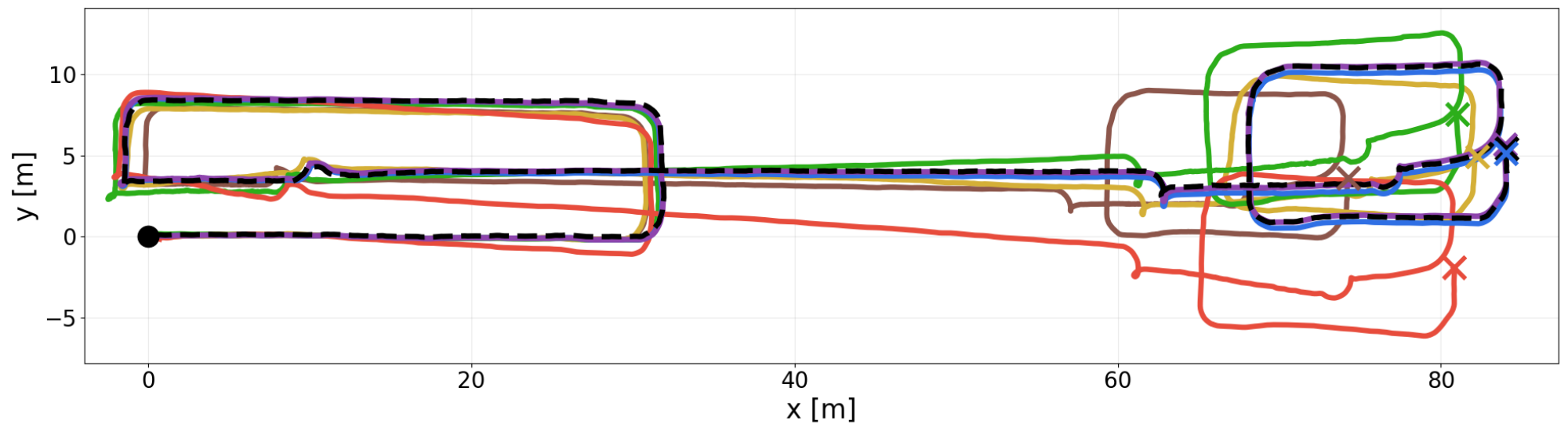}
    \vspace{-2mm}
    \caption{Trajectory comparison with \ac{SOTA} methods on \texttt{Downstair}.}
    \label{fig:downstair}
    \vspace{-6mm}
\end{figure}

\tabref{tab:outdoor} summarizes results in challenging outdoor environments.
Multi-radar RIO and kinematics-based KIO methods complete all sequences, but with limited accuracy. In contrast, stereo VIO performs better in well-lit scenes, but fails under low-light conditions. 
This highlights the limitations of single-modality sensing in challenging environments.
We further compare against RVIO, VKIO, and RKIO methods.
For fair comparison, RAGNAROK-RVI uses the same single-radar monocular visual setup as RVIO baselines. VKIO methods use OKVIS2-X as the stereo visual-inertial odometry, while RKIO methods are configured with multiple radars as in our setup. 
The proposed variants outperform their corresponding baselines on most sequences. RAGNAROK-Odom achieves an average \ac{APE}$_t$ of 0.481, while RAGNAROK-SLAM further supports submapping and achieves the best average \ac{APE}$_t$ of 0.387. Additionally, to evaluate the contribution of our radar fusion strategy, we report results for loosely coupled radar-aided variants of Pronto and MUSE.
Qualitative reconstruction and trajectory results are shown in \figref{fig:mountain} and \figref{fig:overpass}, respectively.
Furthermore, \tabref{tab:indoor} summarizes indoor and hybrid results, while qualitative trajectories are shown in \figref{fig:downstair}.

On visually challenging sequences affected by low illumination or glare, RAGNAROK-RKI performs comparably to or slightly better than RAGNAROK-Odom, indicating that visual measurements do not always improve accuracy. Nevertheless, the exteroceptive pipeline enables loop closure and mapping for downstream tasks such as navigation, while retaining fast processing as detailed in Section~\ref{subsec:runtime}. Moreover, clear gains are observed when informative visual cues are available, as in \texttt{Mountain-Long} and \texttt{Mountain-Short}.

\input{tab/outdoor}

\subsection{Performance on Co-RaL and GaRLILEO Datasets}
We evaluate the generalization ability of the proposed method on the Co-RaL and GaRLILEO datasets. Since both datasets provide radar, IMU, and encoder measurements without visual data, we compare RKIO methods with RAGNAROK-RKI. We also conduct ablation studies to analyze the effects of the following modules: R.C. (rolling-contact modeling) and K.A.C. (kinematics-aware covariance). As shown in \tabref{tab:coral}, RAGNAROK-RKI outperforms existing methods on the Co-RaL dataset, which includes challenging stair and hiking trail sequences. The ablation results further show that both R.C. and K.A.C. improve overall accuracy. Similarly, \tabref{tab:garlileo} shows that RAGNAROK-RKI achieves the best performance, including \texttt{MoCap} sequences designed to induce leg failures through a slippery zone with a backward-moving floor and a cushion zone.

\subsection{Ablation Study}
\begin{figure}[!t]
    \vspace{-0.5mm}
    \centering
    \includegraphics[trim= 0.0cm 0.2cm 0.0cm 0.2cm, clip,width=1.0\columnwidth]{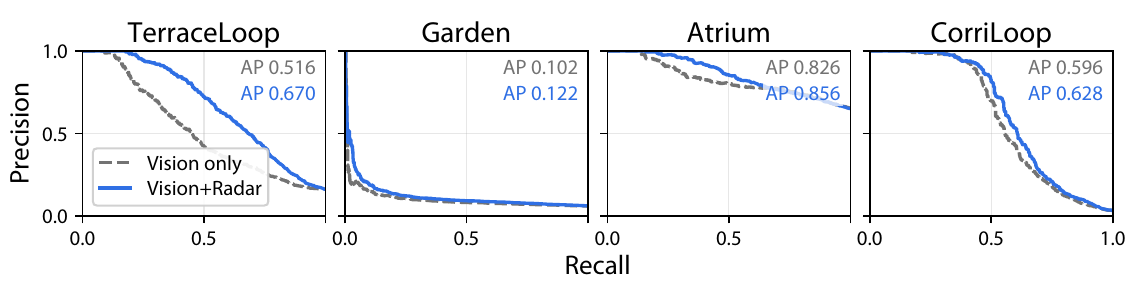}
    \vspace{-7mm}
    \caption{Precision–recall curves for sequences with multiple loops.}
    \label{fig:pr}
    \vspace{-3.5mm}
\end{figure}

We compare RAGNAROK-Odom variants with different combinations of key modules:
\textbf{S} (Slip-aware leg velocity bias), \textbf{I} (Image enhancement), \textbf{G} (Gravity-aligned BRISK), \textbf{M} (Motion-inconsistent landmark rejection), \textbf{A} (Adaptive weighting), \textbf{D} (Dual-radar), and \textbf{O} (Online extrinsic calibration).
As shown in \tabref{tab:ablation}, \textbf{S}, \textbf{G}, \textbf{M}, and \textbf{O} reduce \ac{APE}$_t$ by $4.5\%$, $9.2\%$, $24.3\%$, and $31.8\%$ on \texttt{Overpass}, respectively, where stair-induced oscillations are severe. Their effects are limited on \texttt{TerraceLoop} and \texttt{Atrium}, where body motion is comparatively stable.
In contrast, \textbf{I} and \textbf{A} substantially improve performance across all visually degraded sequences, reducing the average \ac{APE}$_t$ by $59.7\%$ and $87.7\%$, respectively.
Furthermore, \figref{fig:pr} shows that radar-enhanced place recognition improves Average Precision (AP).

\subsection{Runtime Analysis}
\label{subsec:runtime}
\begin{figure}[!t]
    \centering
    \includegraphics[trim= 0.0cm 0.0cm 0.0cm 0.2cm, clip,width=1.0\columnwidth]{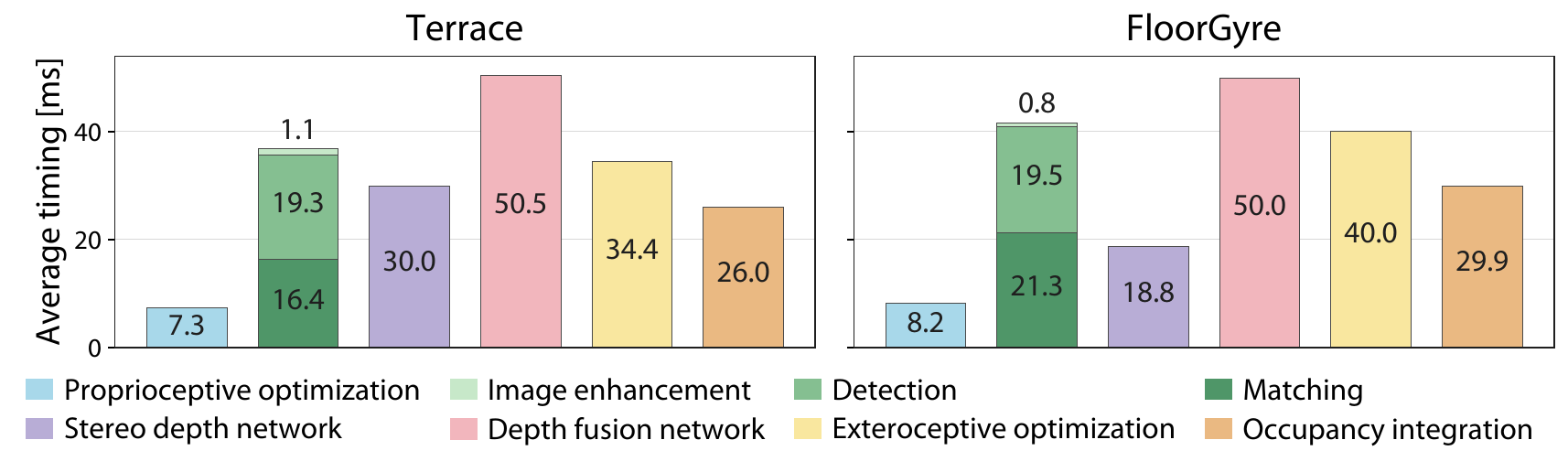}
    \vspace{-7mm}
    \caption{Average runtime of RAGNAROK-SLAM in ms per frame.}
    \label{fig:runtime}
    \vspace{-6mm}
\end{figure}

\figref{fig:runtime} reports the average runtime of each component. All timings were measured on an Intel i7-14700 CPU and an NVIDIA RTX 4080 16 GB GPU. Our multi-threaded design decouples individual modules; therefore, their runtimes are not fully sequential and should not be directly summed.
With sliding-window marginalization, proprioceptive optimization requires 7.3\,ms and 8.2\,ms on \texttt{Terrace} and \texttt{FloorGyre}, respectively, while image enhancement requires only 1.1\,ms and 0.8\,ms, respectively, as \ac{BO} is triggered only by significant illumination changes. On \texttt{Terrace}, asynchronous loop-closure optimization averages 110.7\,ms per closure without blocking real-time estimation, and the overall per-frame runtime of RAGNAROK-SLAM averages 49.5\,ms. These results demonstrate that the system can sustain real-time processing at a camera input rate of 15\,Hz, with only one frame dropped due to transient queue saturation. Although the corresponding end-to-end output latency is 105.5\,ms, the sustainable frame-processing rate is governed by the per-frame runtime rather than by this latency, since the frontend proceeds with the next frame while frame $k$ is still being optimized. On \texttt{FloorGyre}, loop-closure optimization averages 589.3\,ms per closure, and the average runtime increases to 64.8\,ms per frame; however, we note that configurable parameters provide an accuracy-runtime trade-off, enabling faster operation at a small cost in accuracy.
\input{tab/coral}

\input{tab/garlileo}

\input{tab/ablation}

\input{tab/indoor}

%% file: tab/dataset.tex
\begin{table}[!t]
\caption{Comparison with Existing Datasets}
\vspace{-2mm}
\label{table:dataset}
\resizebox{\columnwidth}{!}{%
\begin{tabular}{c|cccccccccc} \toprule \rule{0pt}{2ex}
Dataset & 
Platform &
Radar & 
Camera& 
LiDAR & 
IMU & 
GT & 
Scenarios \\ 
\midrule

\multirow{-1}{*}{RRxIO \cite{doer2021radar}} &
  UAV &
  1 $\times$ SoC &
  1 $\times$ Mono &
  - &
  \checkmark &  {\begin{tabular}[c]{@{}c@{}}Vicon/ \\VISLAM \end{tabular}} &
  {\begin{tabular}[c]{@{}c@{}} Indoor, Outdoor, \\ Dark \end{tabular}}
 \\
  
  



\multirow{-1}{*}{HeRCULES \cite{kim2025hercules}} &
  Car &
  {\begin{tabular}[c]{@{}c@{}} 1 $\times$ 4D, \\  1 $\times$ Spinning\end{tabular}} &
  2 $\times$ Mono &
  \checkmark &
  \checkmark &
  RTK &
  {\begin{tabular}[c]{@{}c@{}} Outdoor, Dark, \\ Rain, Snow \end{tabular}}
 \\
  
\multirow{-1}{*}{Co-RaL \cite{jung2024co}} &
  {\begin{tabular}[c]{@{}c@{}} Legged \\ Robot\end{tabular}}&
  1 $\times$ SoC &
  - &
  \checkmark &
  \checkmark &
  LIO &
  {\begin{tabular}[c]{@{}c@{}} Indoor, Outdoor, \\ Hybrid \end{tabular}}
 \\
 
\multirow{-1}{*}{GaRLILEO \cite{noh2026garlileo}} &
  {\begin{tabular}[c]{@{}c@{}} Legged \\ Robot\end{tabular}}&
  1 $\times$ SoC &
  - &
  \checkmark &
  \checkmark &
  TLS &
  {\begin{tabular}[c]{@{}c@{}} Indoor, Outdoor, \\ Hybrid \end{tabular}}
 \\
\multirow{-1}{*}{GrandTour \cite{tuna2026grandtour}} &
  {\begin{tabular}[c]{@{}c@{}} Legged \\ Robot\end{tabular}}&
  - &
  {\begin{tabular}[c]{@{}c@{}} 7 $\times$ Stereo, \\ 8 $\times$ Mono\end{tabular}}  &
  \checkmark &
  \checkmark & {\begin{tabular}[c]{@{}c@{}} RTS/ \\ GNSS \end{tabular}} &
  {\begin{tabular}[c]{@{}c@{}} Indoor, Outdoor, \\ Dim, Dark, Snow \end{tabular}} 
 \\

  \midrule

  \multirow{-1}{*}{\textbf{RAGNAROK}} &
  {\begin{tabular}[c]{@{}c@{}}\textbf{Legged} \\ \textbf{Robot}\end{tabular}}&
  \textbf{2} \textbf{$\times$} \textbf{SoC} &
  {\begin{tabular}[c]{@{}c@{}} \textbf{1} $\times$ \textbf{Stereo,} \\ \textbf{2} $\times$ \textbf{Mono}\end{tabular}} &
  \textbf{\checkmark} &
  \textbf{\checkmark} &  

  \textbf{TLS} &
  {\begin{tabular}[c]{@{}c@{}} Indoor, Outdoor, Hybrid,\\Dark, Dim, Glare \end{tabular}}  \\
\bottomrule
\end{tabular}%
}
\vspace{-5.5mm}
\end{table}

%% file: tab/sequences.tex
\begin{table}[!t]
\caption{Overview of Sequences}
\vspace{-2mm}
\label{tab:sequence}
\centering
\resizebox{\columnwidth}{!}{
\begin{tabular}{c|cc|cccc} 
\toprule
Sequence & Outdoor & Indoor & Length (m) & Elevation & Self-similarity & \# of Loops\\ 
\midrule

Terrace  & Dark &-& 68.38 & - & -  & 1\\ 
TerraceLoop  & Dark &-& 134.47 & - & -  & 2\\ 
Garden & Dark & -& 162.54  & \checkmark & - & 2\\
Quad & Dim &-& 438.28  & \checkmark &  - & 1\\
Overpass  & Glare &-& 160.10 & \checkmark & -  & 1\\ 
Mountain-Long  & \checkmark &-& 207.59 & \checkmark & - & 1\\
Mountain-Short  & \checkmark &-& 170.14 & \checkmark & - & 1\\

\midrule

Atrium & - & Dark & 68.98  & - & -  & 2\\
FloorGyre & - & Dark & 77.10  & - &  - & -\\
Upstair & -  & Dim & 96.65 & \checkmark &  \checkmark  & - \\ 
Downstair & -  & \checkmark & 216.23 & \checkmark & -  & 1\\ 
CorriLoop & -  & \checkmark & 205.38 & - &  \checkmark  & 2\\
\midrule

 SlopeStair & Glare & \checkmark & 137.25 & \checkmark & -  & 1\\ 
 Tunnel & Dark & Dim & 208.51  & - &   \checkmark & - \\ 


  
\bottomrule
\end{tabular}
}
\vspace{-6mm}
\end{table}

%% file: tab/outdoor.tex
\newcommand{\rankone}[1]{\cellcolor[HTML]{CBE4CF}\textbf{#1}}
\newcommand{\ranktwo}[1]{\cellcolor[HTML]{E5EDC3}#1}
\newcommand{\rankthree}[1]{\cellcolor[HTML]{FBF8CC}#1}

\begin{table*}[!t]
\centering
\caption{Evaluation Results on Outdoor Sequences (R: Radar, V: Visual, K: Kinematic, I: Inertial, \protect\colorbox[HTML]{CBE4CF}{\textbf{1st}}, \protect\colorbox[HTML]{E5EDC3}{2nd}, \protect\colorbox[HTML]{FBF8CC}{3rd})}
\vspace{-2mm}
\label{tab:outdoor}
\resizebox{\textwidth}{!}{%
\begin{tabular}{cccc|l|cc|cc|cc|cc|cc|cc|cc|cc}
\toprule
 \multicolumn{4}{c|}{Modality} & \multirow{2}{*}{Methods} & 
  \multicolumn{2}{c|}{\texttt{Terrace}} &
  \multicolumn{2}{c|}{\texttt{TerraceLoop}} &
  \multicolumn{2}{c|}{\texttt{Garden}} &
  \multicolumn{2}{c|}{\texttt{Quad}} &
  \multicolumn{2}{c|}{\texttt{Overpass}} &
  \multicolumn{2}{c|}{\texttt{Mountain-Long}} &
  \multicolumn{2}{c|}{\texttt{Mountain-Short}} &
  \multicolumn{2}{c}{Average}
 \\
R & V & K & I &  & 
   APE$_t$ & APE$_r$ &
   APE$_t$ & APE$_r$ &
   APE$_t$ & APE$_r$ &
   APE$_t$ & APE$_r$ &
   APE$_t$ & APE$_r$ &
   APE$_t$ & APE$_r$ &
   APE$_t$ & APE$_r$ &
   APE$_t$ & APE$_r$ \\
\midrule

\checkmark&&&\checkmark
  & x-RIO~\cite{doer2021x} 
  & 1.655 & 7.384 
  & 2.829 & 8.555 
  & 4.316 & 18.500 
  & 16.595 & 21.122 
  & 3.222 & 8.068 
  & 10.096 & 10.139 
  & 9.691 & 9.421 
  & 6.915 & 11.884 \\
\midrule

&\multirow{2}{*}{\checkmark}&& \multirow{2}{*}{\checkmark}
  & OpenVINS~\cite{geneva2020openvins} 
  & - & - 
  & - & - 
  & - & - 
  & 8.275 & 2.970 
  & 0.591 & \rankone{2.045} 
  & 0.406 & \rankone{0.754} 
  & 0.564 & 0.889 
  & - & - \\
&&&
  & OKVIS2-X~\cite{boche2025okvis2} 
  & 0.474 & 2.596 
  & 0.568 & 3.073 
  & 6.626 & 10.046 
  & - & - 
  & 1.166 & 3.447 
  & 0.311 & 1.196 
  & 0.465 & 1.014 
  & - & - \\ 
\midrule

&&\multirow{2}{*}{\checkmark} & \multirow{2}{*}{\checkmark}
  & DRIFT~\cite{lin2023proprioceptive} 
  & 1.767 & 7.466 
  & 2.942 & 6.116 
  & 4.676 & 18.563 
  & 13.150 & 8.879 
  & 4.691 & 3.569 
  & 7.661 & 4.805 
  & 6.473 & 5.964 
  & 5.909 & 7.909 \\
&&& 
  & Holistic~\cite{nubert2026holistic} 
  & 2.077 & 13.082 
  & 6.152 & 15.696 
  & 10.225 & 32.851 
  & 27.420 & 25.328 
  & 6.937 & 17.755 
  & 2.101 & 8.072 
  & 13.689 & 21.850 
  & 9.800 & 19.233 \\ 
\midrule

\multirow{3}{*}{\checkmark}&\multirow{3}{*}{\checkmark}&& \multirow{3}{*}{\checkmark}
  & RRxIO~\cite{doer2021radar} 
  & 0.369 & 2.370
  & 1.176 & 1.998
  & - & -
  & - & -
  & 5.779 & 3.688
  & 3.506 & 1.885 
  & 5.085 & 1.650 
  & - & - \\
&&&
  & RadVIO~\cite{nissov2026tightly} 
  & 5.941 & 19.316
  & 5.077 & 5.889
  & - & -
  & - & -
  & 2.111	& 2.261 
  & 2.836	& 3.513	
  & 3.545	& 1.971
  & - & - \\
\cmidrule{5-21}
&&&
  & \cellcolor[HTML]{f3f7fc}RAGNAROK-RVI 
  & 0.499 & 2.988 
  & 0.911 & 3.143 
  & - & - 
  & - & - 
  & 6.497 & 9.154 
  & 0.532 & \rankthree{0.872} 
  & 0.756 & 1.126 
  & - & - \\
\midrule

&\multirow{3}{*}{\checkmark}&\multirow{3}{*}{\checkmark}& \multirow{3}{*}{\checkmark}
  & Pronto*~\cite{camurri2020pronto} 
  & 0.709 & 1.905 
  & 0.878 & 1.920 
  & - & - 
  & - & - 
  & 2.703 & 2.912 
  & 6.999 & 3.073 
  & 5.546 & 3.104 
  & - & - \\
&&&
  & MUSE*~\cite{nistico2025muse} 
  & 0.626 & 2.354 
  & 0.983 & 2.650 
  & 2.984 & 5.805 
  & - & - 
  & 0.749 & \ranktwo{2.136} 
  & 0.913 & 1.437 
  & 0.445 & \ranktwo{0.797} 
  & - & - \\
\cmidrule{5-21}
&&&
  & \cellcolor[HTML]{f3f7fc}RAGNAROK-VKI 
  & 0.224 & 2.001 
  & 0.344 & 1.723 
  & 0.625 & 2.320 
  & 4.354 & 4.180 
  & 5.401 & 2.246 
  & \ranktwo{0.222} & 0.958 
  & 0.390 & 0.981 
  & 1.651 & \rankthree{2.058} \\
\midrule

\multirow{3}{*}{\checkmark}&&\multirow{3}{*}{\checkmark}& \multirow{3}{*}{\checkmark}
  & Co-RaL~\cite{jung2024co} 
  & 0.485 & 2.804 
  & 1.048 & 3.177 
  & 1.216 & 5.568 
  & 4.180 & 7.182 
  & 0.848 & 3.702 
  & 5.019 & 4.591 
  & 5.571 & 5.932 
  & 2.624 & 4.708 \\
&&&
  & GaRLILEO~\cite{noh2026garlileo} 
  & 0.363 & 2.290 
  & 0.459 & 1.987 
  & 1.165 & 5.925 
  & \rankthree{1.810} & 2.585 
  & 1.096 & 2.625 
  & 2.854 & 3.044 
  & 2.367 & 3.165 
  & 1.445 & 3.089 \\
\cmidrule{5-21}
&&&
  & \cellcolor[HTML]{f3f7fc}RAGNAROK-RKI 
  & \rankthree{0.143} & \rankone{1.671} 
  & \rankthree{0.226} & \rankthree{1.543} 
  & \rankone{0.223} & \ranktwo{1.234} 
  & 1.889 & \rankthree{2.109} 
  & \ranktwo{0.439} & \rankthree{2.175} 
  & 2.927 & 3.638 
  & 1.704 & 3.145 
  & \rankthree{1.079} & 2.216 \\
\midrule

\multirow{4}{*}{\checkmark}&\multirow{4}{*}{\checkmark}&\multirow{4}{*}{\checkmark}& \multirow{4}{*}{\checkmark}
   & Pronto*~\cite{camurri2020pronto} + Radar
   & 0.703
   & 1.861
   & 0.864
   & 1.895	
   & 1.131	
   & 3.603
   & -
   & -
   & 2.666	
   & 2.851
   & 7.055	
   & 4.356
   & 5.546	
   & 3.093
   & -
   & -
   \\
   
&&&
   & MUSE*~\cite{nistico2025muse} + Radar
   & 0.671	
   & 2.251	
   & 0.651	
   & 2.550	
   & 2.640	
   & 7.337
   & -
   & -
   & 0.998	
   & 2.820
   & 0.553	
   & 1.212	
   & \ranktwo{0.236}
   & 0.806
   & -
   & -
   \\
   
\cmidrule{5-21}
&&&
   & \cellcolor[HTML]{f3f7fc}RAGNAROK-Odom
   & \ranktwo{0.140} 
   & \rankthree{1.710} 
   & \ranktwo{0.219} 
   & \ranktwo{1.542} 
   & \rankthree{0.227} 
   & \rankthree{1.358} 
   & \ranktwo{1.660} 
   & \rankone{2.079} 
   & \rankthree{0.514} 
   & 2.309 
   & \rankthree{0.283} 
   & 1.050 
   & \rankthree{0.326}
   & \rankthree{0.805} 
   & \ranktwo{0.481}
   & \ranktwo{1.550} \\
&&&
   & \cellcolor[HTML]{f3f7fc}RAGNAROK-SLAM
   & \rankone{0.135} & \ranktwo{1.698} 
   & \rankone{0.195} & \rankone{1.504} 
   & \ranktwo{0.226} & \rankone{1.164} 
   & \rankone{1.649} & \ranktwo{2.100} 
   & \rankone{0.294} & 2.315 
   & \rankone{0.097} & \ranktwo{0.779} 
   & \rankone{0.114} & \rankone{0.638} 
   & \rankone{0.387} & \rankone{1.457} \\
\bottomrule
\multicolumn{20}{l}{\hspace{0.8em}
All quantitative results are reported as the median over 5 runs.
}\\
\multicolumn{20}{l}{* For multi-sensor fusion in Pronto and MUSE, visual-inertial odometry was obtained using OKVIS2-X in our experiments.}
\end{tabular}%
}

\vspace{-5mm}
\end{table*}

%% file: tab/coral.tex
\providecommand{\rankone}[1]{\cellcolor[HTML]{CBE4CF}\textbf{#1}}
\providecommand{\ranktwo}[1]{\cellcolor[HTML]{E5EDC3}#1}

\newcolumntype{C}[1]{>{\centering\arraybackslash}m{#1}}
\begin{table}[!t]
\centering
\caption{Evaluation Results on Co-RaL Dataset (\ac{APE}$_t$)}
\vspace{-2mm}
\label{tab:coral}
\resizebox{\columnwidth}{!}{
\begin{tabular}{l|C{1.6cm}|C{1.6cm}|C{1.4cm}|C{1.4cm}|c}
\toprule
Methods
& \texttt{Building}
& \texttt{Garage}
& \texttt{Stair}
& \texttt{Trail}
& Average
\\
\midrule
Co-RaL~\cite{jung2024co} 
& 4.922 
& 1.494
& 1.562
& 2.586
& 2.641
\\

GaRLILEO~\cite{noh2026garlileo} 
& 4.017
& 1.367
& 3.040
& 2.888
& 2.828
\\
\midrule

\cellcolor[HTML]{f3f7fc}RAGNAROK-RKI (w/o R.C.)
& \ranktwo{3.565}
& \ranktwo{0.890}
& \ranktwo{0.856} 
& 0.993
& 1.576
\\

\cellcolor[HTML]{f3f7fc}RAGNAROK-RKI (w/o K.A.C.)
& 3.610
& \rankone{0.879}
& \ranktwo{0.856} 
& \rankone{0.951}
& \ranktwo{1.574}
\\

\cellcolor[HTML]{f3f7fc}RAGNAROK-RKI 
& \rankone{3.484} 
& \ranktwo{0.890}
& \rankone{0.826} 
& \ranktwo{0.962}
& \rankone{1.541}
\\
\bottomrule
\end{tabular}
}
\vspace{-3mm}
\end{table}

%% file: tab/garlileo.tex
\providecommand{\rankone}[1]{\cellcolor[HTML]{CBE4CF}\textbf{#1}}
\providecommand{\ranktwo}[1]{\cellcolor[HTML]{E5EDC3}#1}

\newcolumntype{C}[1]{>{\centering\arraybackslash}m{#1}}
\begin{table}[!t]
\centering
\caption{Evaluation Results on GaRLILEO Dataset (\ac{APE}$_t$)}
\vspace{-2mm}
\label{tab:garlileo}
\resizebox{\columnwidth}{!}{
\begin{tabular}{l|C{1.6cm}|C{1.6cm}|C{1.4cm}|C{1.4cm}|c}
\toprule
Methods
& \texttt{BridgeLoop}
& \texttt{SlopeStair}
& \texttt{MoCap-E}
& \texttt{MoCap-H}
& Average
\\ \midrule

Co-RaL~\cite{jung2024co} 
& 2.182
& 4.696
& -
& 1.549
& -
\\

GaRLILEO~\cite{noh2026garlileo} 
& 0.345
& 1.132
& 0.356
& \ranktwo{0.467}
& 0.575
\\
\midrule

\cellcolor[HTML]{f3f7fc}RAGNAROK-RKI (w/o R.C.)
& 0.408
& \ranktwo{0.946}
& \ranktwo{0.346} 
& \rankone{0.365} 
& \ranktwo{0.516}
\\

\cellcolor[HTML]{f3f7fc}RAGNAROK-RKI (w/o K.A.C.)
& \rankone{0.326}
& 1.089
& \ranktwo{0.346} 
& \rankone{0.365} 
& 0.532
\\

\cellcolor[HTML]{f3f7fc}RAGNAROK-RKI 
& \ranktwo{0.338}
& \rankone{0.905}
& \rankone{0.345} 
& \rankone{0.365} 
& \rankone{0.488}
\\
\bottomrule
\end{tabular}
}
\vspace{-3mm}
\end{table}

%% file: tab/ablation.tex
\providecommand{\rankone}[1]{\cellcolor[HTML]{CBE4CF}\textbf{#1}}
\providecommand{\ranktwo}[1]{\cellcolor[HTML]{E5EDC3}#1}

\begin{table}[!t]
\centering
\caption{Ablation Study on RAGNAROK Components}
\vspace{-2mm}
\label{tab:ablation}
\resizebox{\columnwidth}{!}{
\begin{tabular}{ccccccc|cc|cc|cc|cc}
\toprule
\multirow{2}{*}{\textbf{S}} &
\multirow{2}{*}{\textbf{I}} &
\multirow{2}{*}{\textbf{G}} &
\multirow{2}{*}{\textbf{M}} &
\multirow{2}{*}{\textbf{A}} &
\multirow{2}{*}{\textbf{D}} &
\multirow{2}{*}{\textbf{O}} &
\multicolumn{2}{c|}{\texttt{TerraceLoop}} &
\multicolumn{2}{c|}{\texttt{Overpass}} &
\multicolumn{2}{c|}{\texttt{Atrium}} &
\multicolumn{2}{c}{Average}
\\
& & & & & & &
APE$_t$ & APE$_r$ &
APE$_t$ & APE$_r$ &
APE$_t$ & APE$_r$ &
APE$_t$ & APE$_r$
\\
\midrule

$\times$ &
\checkmark &
\checkmark &
\checkmark &
\checkmark &
\checkmark &
\checkmark
& \ranktwo{0.223}
& 1.619
& \ranktwo{0.538}
& \ranktwo{2.317}
& \ranktwo{0.084}
& 1.656
& \ranktwo{0.282}
& 1.864
\\

\checkmark &
$\times$ &
\checkmark &
\checkmark &
\checkmark &
\checkmark &
\checkmark
& 0.753
& 1.952
& 0.806
& 3.557
& 0.466
& 2.314
& 0.675
& 2.608
\\

\checkmark &
\checkmark &
$\times$ &
\checkmark &
\checkmark &
\checkmark &
\checkmark
& 0.254
& 1.598
& 0.566
& 2.513
& 0.087
& \rankone{1.582}
& 0.302
& 1.898
\\

\checkmark &
\checkmark &
\checkmark &
$\times$ &
\checkmark &
\checkmark &
\checkmark
& \ranktwo{0.223}
& 1.588
& 0.679
& 2.729
& 0.085
& 1.623
& 0.329
& 1.980
\\

\checkmark &
\checkmark &
\checkmark &
\checkmark &
$\times$ &
\checkmark &
\checkmark
& 0.785
& 3.549
& 1.657
& 5.142
& 4.217
& 16.370
& 2.220
& 8.354
\\

\checkmark &
\checkmark &
\checkmark &
\checkmark &
\checkmark &
$\times$ &
\checkmark
& 0.236
& \ranktwo{1.557}
& 0.549
& 2.391
& \rankone{0.078}
& 1.604
& 0.288
& \ranktwo{1.851}
\\

\checkmark &
\checkmark &
\checkmark &
\checkmark &
\checkmark &
\checkmark &
$\times$
& \rankone{0.219}
& 1.558
& 0.754
& 2.643
& 0.085
& 1.604
& 0.353
& 1.935
\\

\midrule

\checkmark &
\checkmark &
\checkmark &
\checkmark &
\checkmark &
\checkmark &
\checkmark
& \rankone{0.219}
& \rankone{1.542}
& \rankone{0.514}
& \rankone{2.309}
& \ranktwo{0.084}
& \ranktwo{1.599}
& \rankone{0.272}
& \rankone{1.817}
\\

\bottomrule
\end{tabular}
}
\vspace{-3mm}
\end{table}

%% file: tab/indoor.tex
\providecommand{\rankone}[1]{\cellcolor[HTML]{CBE4CF}\textbf{#1}}
\providecommand{\ranktwo}[1]{\cellcolor[HTML]{E5EDC3}#1}
\providecommand{\rankthree}[1]{\cellcolor[HTML]{FBF8CC}#1}

\begin{table*}[!t]
\centering
\caption{
Evaluation Results on Indoor and Hybrid Sequences (R: Radar, V: Visual, K: Kinematic, I: Inertial, \protect\colorbox[HTML]{CBE4CF}{\textbf{1st}}, \protect\colorbox[HTML]{E5EDC3}{2nd}, \protect\colorbox[HTML]{FBF8CC}{3rd})}

\vspace{-2mm}
\label{tab:indoor}
\resizebox{\textwidth}{!}{%
\begin{tabular}{cccc|l|cc|cc|cc|cc|cc|cc|cc|cc}
\toprule
 \multicolumn{4}{c|}{Modality} & \multirow{2}{*}{Methods} & 
  \multicolumn{2}{c|}{\texttt{Upstair}} &
  \multicolumn{2}{c|}{\texttt{Downstair}} &
  \multicolumn{2}{c|}{\texttt{FloorGyre}} &
  \multicolumn{2}{c|}{\texttt{CorriLoop}} &
  \multicolumn{2}{c|}{\texttt{Atrium}} &
  \multicolumn{2}{c|}{\texttt{SlopeStair}} &
  \multicolumn{2}{c|}{\texttt{Tunnel}} &
  \multicolumn{2}{c}{Average}
 \\
R & V & K & I &  & 
   APE$_t$ & APE$_r$ &
   APE$_t$ & APE$_r$ &
   APE$_t$ & APE$_r$ &
   APE$_t$ & APE$_r$ &
   APE$_t$ & APE$_r$ &
   APE$_t$ & APE$_r$ &
   APE$_t$ & APE$_r$ &
   APE$_t$ & APE$_r$ \\
\midrule

\checkmark&&&\checkmark
  & x-RIO~\cite{doer2021x} 
  & 2.020 & 7.873 
  & 1.755 & 6.715 
  & 1.123 & 6.587 
  & 2.535 & 7.361 
  & 1.562 & 6.934 
  & 1.497 & 7.277
  & 8.181 & 29.785
  & 2.668 & 10.362 \\
\midrule

&\multirow{2}{*}{\checkmark}&& \multirow{2}{*}{\checkmark}
  & OpenVINS~\cite{geneva2020openvins} 
  & \ranktwo{0.351} & \rankone{1.433} 
  & 0.701 & 1.873
  & - & - 
  & 0.631 & \rankone{1.293} 
  & - & - 
  & 0.410 & \rankthree{1.352}
  & 3.611 & 30.145
  & - & - \\
&&&
  & OKVIS2-X~\cite{boche2025okvis2} 
  & 1.008 & 2.957 
  & 0.539 & \rankone{1.661} 
  & 0.872 & 3.536 
  & 0.427 & \rankthree{1.500} 
  & - & - 
  & \rankone{0.261} & \rankone{1.252}
  & 2.550 & 5.260
  & - & - \\ 
\midrule

&&\multirow{2}{*}{\checkmark} & \multirow{2}{*}{\checkmark}
  & DRIFT~\cite{lin2023proprioceptive} 
  & 2.139 & 6.010 
  & 3.870 & 10.425 
  & 2.016 & 8.602 
  & 5.656 & 9.392 
  & 2.204 & 8.220 
  & 3.099 & 6.396
  & 6.454 & 42.345
  & 3.634 & 13.056 \\
&&& 
  & Holistic~\cite{nubert2026holistic} 
  & 2.443 & 13.713 
  & 10.834 & 27.907 
  & 5.110 & 27.343 
  & 10.834 & 27.907 
  & 3.327 & 25.615 
  & 6.170 & 25.278
  & 14.001 & 23.840
  & 7.531 & 24.515 \\ 
\midrule

\multirow{3}{*}{\checkmark}&\multirow{3}{*}{\checkmark}&& \multirow{3}{*}{\checkmark}
  & RRxIO~\cite{doer2021radar} 
  & 1.398 & 4.405
  & 1.714	& 2.187
  & 0.776 & 4.905
  & 1.826 & 3.058
  & 0.742 & 3.005
  & 1.573 & 2.527
  & - & -
  & - & - \\
&&&
  & RadVIO~\cite{nissov2026tightly} 
  & 2.071 & 5.141
  & -	& -
  & 1.710	& 7.099
  & 12.945 & 12.209
  & 5.342	& 13.783
  & 1.335 & 5.363
  & - & -
  & - & - \\   
\cmidrule{5-21}
&&&
  & \cellcolor[HTML]{f3f7fc}RAGNAROK-RVI 
  & 1.036 & 2.981 
  & 1.537 & \rankone{1.661} 
  & 0.932 & 3.740 
  & 1.043 & \ranktwo{1.441} 
  & 16.795 & 19.814 
  & 0.687 & \ranktwo{1.302}
  & 9.067 & 14.634
  & 4.442 & 6.510 \\
\midrule

&\multirow{3}{*}{\checkmark}&\multirow{3}{*}{\checkmark}& \multirow{3}{*}{\checkmark}
  & Pronto*~\cite{camurri2020pronto} 
  & 0.998 & 2.469 
  & 4.256 & \rankthree{1.713} 
  & 0.531 & 2.468 
  & 1.323 & 2.041 
  & 0.887 & 2.478 
  & 1.619 & 2.422
  & 2.303 & 9.956
  & 1.702 & 3.364 \\
&&&
  & MUSE*~\cite{nistico2025muse} 
  & 0.782 & 2.639 
  & 0.848 & 2.107 
  & 0.605 & 3.188
  & 0.396 & 1.618 
  & 27.980 & 46.512 
  & 1.479 & 2.201
  & 1.959 & 11.437
  & 4.864 & 9.957 \\
\cmidrule{5-21}
&&&
  & \cellcolor[HTML]{f3f7fc}RAGNAROK-VKI 
  & 0.503 & 2.205 
  & \rankthree{0.359} & 2.984 
  & 0.228 & 2.564 
  & 0.338 & 1.945 
  & 0.232 & 1.848 
  & 2.641 & 5.022
  & 1.392 & 4.992
  & 0.813 & 3.080 \\
\midrule
  
\multirow{3}{*}{\checkmark}&&\multirow{3}{*}{\checkmark}& \multirow{3}{*}{\checkmark}
  & Co-RaL~\cite{jung2024co} 
  & \rankone{0.260} & 2.653 
  & 1.202 & 2.796 
  & 0.745 & 3.031 
  & 0.845 & 3.281 
  & 0.938 & 3.927 
  & 1.207 & 3.901
  & 1.792 & \rankone{3.168}
  & 0.998 & 3.251 \\
&&&
  & GaRLILEO~\cite{noh2026garlileo} 
  & 0.435 & 3.255 
  & 1.098 & 5.378 
  & 0.424 & 3.360 
  & 1.244 & 2.749 
  & 0.533 & 4.281 
  & 1.388 & 5.786
  & 1.167 & 20.041
  & 0.898 & 6.407 \\ 
\cmidrule{5-21}
&&&
  & \cellcolor[HTML]{f3f7fc}RAGNAROK-RKI 
  & 0.399 & \rankthree{2.053} 
  & \rankone{0.231} & 2.284 
  & \rankthree{0.138} & \rankone{2.233} 
  & \rankone{0.261} & 1.692 
  & \rankone{0.077} & \rankone{1.474} 
  & 0.347 & 3.649
  & \ranktwo{0.474} & 5.833
  & \ranktwo{0.275} & \rankthree{2.745} \\
\midrule
  
\multirow{4}{*}{\checkmark}&\multirow{4}{*}{\checkmark}&\multirow{4}{*}{\checkmark}& \multirow{4}{*}{\checkmark}
   & Pronto*~\cite{camurri2020pronto} + Radar
   & 0.954	
   & 2.375	
   & 4.171	
   & \ranktwo{1.688}
   & 0.521	
   & \rankthree{2.418}
   & 1.238	
   & 0.617	
   & 2.556
   & 1.560
   & 2.331
   & 2.218	
   & 7.302
   & 1.611
   & 2.931
   \\
   
&&&
   & MUSE*~\cite{nistico2025muse} + Radar
   & 0.585	
   & 2.412	
   & 0.764	
   & 2.046
   & 0.649	
   & 3.662
   & 0.531	
   & 1.921
   & -
   & -
   & \ranktwo{0.272}
   & 1.389	
   & 1.424	
   & \rankthree{4.427}
   & -
   & -
   \\
   
\cmidrule{5-21}
&&&
   & \cellcolor[HTML]{f3f7fc}RAGNAROK-Odom
   & 0.377 & \ranktwo{1.980} 
   & \ranktwo{0.248} & 2.358 
   & \ranktwo{0.128} & \ranktwo{2.417} 
   & \rankthree{0.279} & 1.806 
   & \ranktwo{0.084} & \ranktwo{1.599} 
   & 0.339 & 3.601
   & \rankthree{0.475} & \ranktwo{4.068}
   & \rankthree{0.276} & \rankone{2.547} \\
&&&
   & \cellcolor[HTML]{f3f7fc}RAGNAROK-SLAM 
   & \rankthree{0.355} & 2.143 
   & \ranktwo{0.248} & 2.392 
   & \rankone{0.121} & 2.425
   & \ranktwo{0.273} & 1.795 
   & \rankthree{0.105} & \rankthree{1.759} 
   & \rankthree{0.284} & 3.070
   & \rankone{0.468} & 4.542
   & \rankone{0.265} & \ranktwo{2.589} \\
\bottomrule
\end{tabular}%
}
\vspace{-5mm}
\end{table*}

%% file: 5_conclusion.tex
\section{Conclusion}
In this letter, we presented a radar-visual-kinematic-inertial \ac{SLAM} framework for robust state estimation in challenging environments. 
It addresses key limitations of existing legged-robot odometry methods, including limited yaw constraints, slip-induced proprioceptive corruption, and perceptual degradation.
Our framework combines a slip- and rolling-contact-aware leg factor, a kinematics-aware multi-radar factor, a degradation-aware visual enhancement, radar-enhanced place recognition, and adaptive multi-sensor weighting within a unified architecture. Extensive experiments demonstrated that RAGNAROK achieves more robust performance than \ac{SOTA} baselines in low-light and high-vibration scenarios.

%% file: 6_acknowledgments.tex
\section{Acknowledgments}
The authors used AI-assisted tools in the refinement of \figref{fig:robot}. Final verification was performed by the authors.